\documentclass[runningheads]{llncs}
\usepackage{graphicx}
\usepackage{tikz}
\usepackage{comment}
\usepackage{amsmath,amssymb} % define this before the line numbering.
\usepackage{color}
\usepackage{ulem}
\usepackage[T1]{fontenc}
\usepackage{multirow}
\usepackage{cite}
\usepackage{wrapfig}

\usepackage[pagebackref,breaklinks,colorlinks]{hyperref}

\usepackage[accsupp]{axessibility}  % Improves PDF readability for those with disabilities.

\newcommand{\sdalle}[1]{\textsc{StoryDALL-E}}
\newcommand{\sgan}[1]{\textsc{StoryGANc}}
\newcommand{\sdallem}[1]{\textsc{mega-StoryDALL-E}}

\begin{document}
% \linenumbers
\pagestyle{headings}
\mainmatter
\def\ECCVSubNumber{8009}  % Insert your submission number here

\title{\sdalle{}: Adapting Pretrained Text-to-Image Transformers for Story Continuation} 
% Replace with your title

%\begin{comment}
% \titlerunning{ECCV-22 submission ID \ECCVSubNumber} 
% \authorrunning{ECCV-22 submission ID \ECCVSubNumber} 
% \author{Anonymous ECCV submission}
% \institute{Paper ID \ECCVSubNumber}
%\end{comment}
%******************


\titlerunning{StoryDALL-E}
\author{Adyasha Maharana \and
Darryl Hannan \and
Mohit Bansal}
\authorrunning{Maharana et al.}
\institute{UNC Chapel Hill, NC 27514, USA\\
\email{\{adyasha,dhannan,mbansal\}@cs.unc.edu}}

%******************
\maketitle

\begin{abstract}
Recent advances in text-to-image synthesis have led to large pretrained transformers with excellent capabilities to generate visualizations from a given text. However, these models are ill-suited for specialized tasks like story visualization, which requires an agent to produce a sequence of images given a corresponding sequence of captions, forming a narrative. Moreover, we find that the story visualization task fails to accommodate generalization to unseen plots and characters in new narratives. Hence, we first propose the task of story continuation, where the generated visual story is conditioned on a source image, allowing for better generalization to narratives with new characters. It is difficult to collect large-scale datasets to train large models for this task from scratch due to the need for continuity and an explicit narrative among the images in a story. Therefore, we propose to leverage the pretrained knowledge of text-to-image synthesis models to overcome the low-resource scenario and improve generation for story continuation. To that end, we enhance or `retro-fit' the pretrained text-to-image synthesis models with task-specific modules for (a) sequential image generation and (b) copying relevant elements from an initial frame. Then, we explore full-model finetuning, as well as prompt-based tuning for parameter-efficient adaptation, of the pre-trained model. We evaluate our approach \sdalle{} on two existing datasets, PororoSV and FlintstonesSV, and introduce a new dataset DiDeMoSV collected from a video-captioning dataset. We also develop a model \sgan{} based on Generative Adversarial Networks (GAN) for story continuation, and compare it with the \sdalle{} model to demonstrate the advantages of our approach. We show that our retro-fitting approach outperforms GAN-based models for story continuation and facilitates copying of visual elements from the source image, thereby improving continuity in the generated visual story. Finally, our analysis suggests that pretrained transformers struggle to comprehend narratives containing several characters and translating them into appropriate imagery. Overall, our work demonstrates that pretrained text-to-image synthesis models can be adapted for complex and low-resource tasks like story continuation. Our results encourage future research into story continuation as well as exploration of the latest,  larger models for the task. Code, data, demo and model card available at \url{https://github.com/adymaharana/storydalle}.

\end{abstract}

\section{Introduction}
\label{sec:intro}

\begin{figure*}[t]
    \centering
    \includegraphics[width=0.96\textwidth]{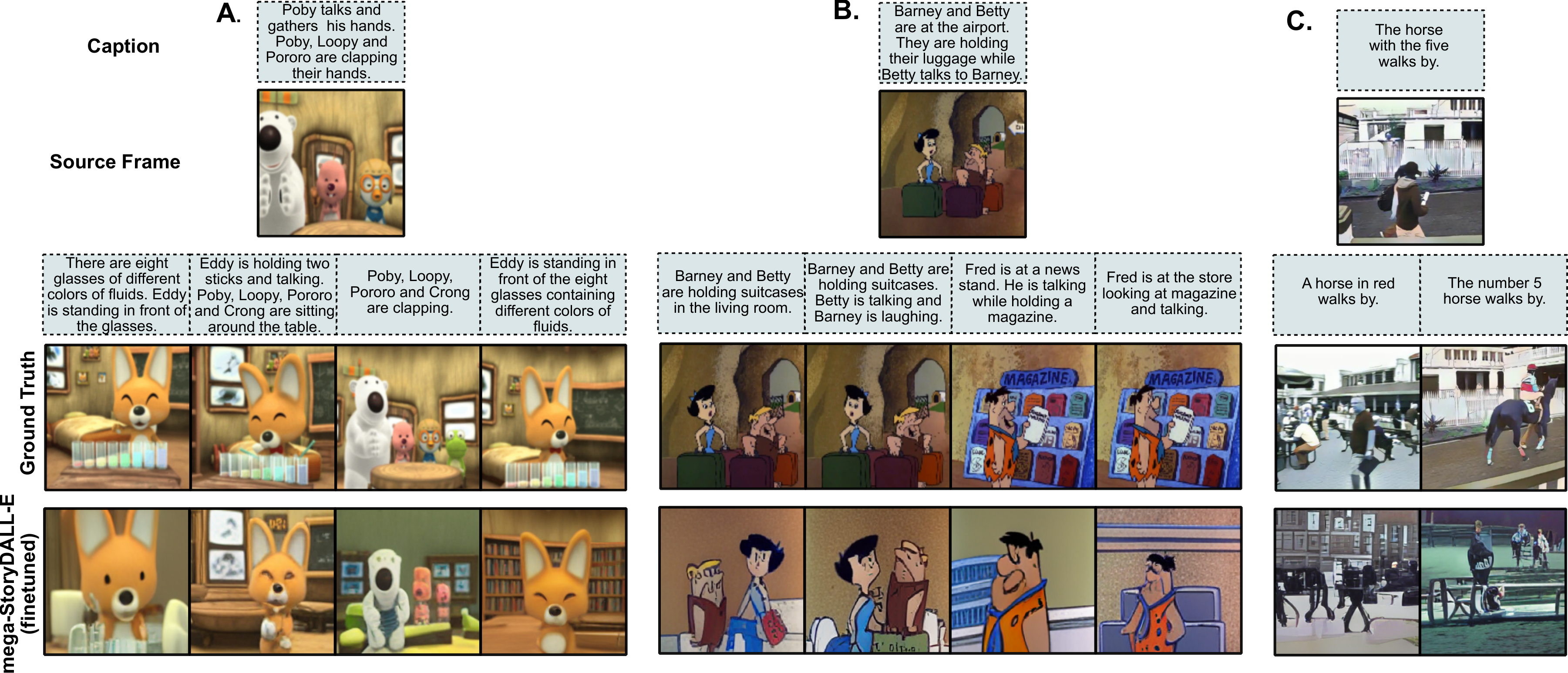}
    \caption{Examples of predictions for (A) PororoSV (B) FlintstonesSV and (C) DiDeMoSV story continuation datasets from the \sdallem{} model. Source frame refers to the initial frame provided as additional input to the model.}
    \label{fig:predictions_mega}
\end{figure*}

Pretrained text-to-image synthesis models like DALL-E \cite{ramesh2021zero} have shown unprecedented ability to convert an input caption into a coherent visualization. Several subsequent approaches have also leveraged powerful multimodal models \cite{radford2021learning, esser2021taming} for creating artistic renditions of input captions \cite{frans2021clipdraw}, demonstrating their potential for democratizing art. However, these models are designed to process only a single, short caption as input. In contrast, many use cases of text-to-image synthesis require models to process long narratives and metaphorical expressions, condition on existing visuals and generate more than one image to capture the meaning of the input text. In the past, multiple works have developed specialized Generative Adversarial Networks (GAN) models such as image-to-image translation \cite{isola2017image}, style transfer \cite{karras2019style} etc. For instance, story visualization models \cite{li2019storygan} convert a sequence of captions into a sequence of images that illustrate the story. However, the recent advent of transformer-based large pretrained models opens up possibilities for leveraging latent knowledge from large-scale pretrained datasets for performing these specialized tasks more effectively, in a paradigm that is similar to finetuning of pretrained language models for performing downstream tasks based on language understanding.
Hence, in this paper, we explore methods to adapt a pretrained text-to-image synthesis model for complex downstream tasks, with a focus on story visualization.

Story visualization is a challenging task that lies at the intersection of image generation and narrative understanding. Given a series of captions, which compose a story, an agent must generate a corresponding sequence of images that depicts the contents of these captions. While prior work in story visualization has discussed potential applications of the task \cite{maharana2021improving, maharana2021integrating, li2019storygan, song2020CPCSV}, the task itself presents some difficulties when being applied to real world settings. The model is limited to the fixed set of characters, settings, and events on which it is trained and has no way of knowing how to depict a new character that appears in a caption during test time; captions do not contain enough information to fully describe the character's appearance. Therefore, in order to generalize to new story elements, the model must have a mechanism for obtaining additional information about how these elements should be visually represented. First, we make story visualization more conducive to these use cases by presenting a new task called `story continuation'. In this task, we provide an initial scene that can be obtained in real world use cases. By including this scene, the model can then copy and adapt elements from it as it generates subsequent images (see Fig.~\ref{fig:predictions_mega}). This has the additional benefit of shifting the focus from text-to-image generation, which is already a task attracting plenty of research, and instead focuses on the narrative structure of a sequence of images, e.g., how an image should change over time to reflect new narrative information in the captions. We introduce a new dataset, DiDeMoSV \cite{hendricks17iccv}, and also convert two existing visualization datasets PororoSV \cite{li2019storygan} and FlintstonesSV \cite{gupta2018imagine} to the story continuation setting. 

Next, in order to adapt a text-to-image synthesis model to this story continuation task, we need to finetune the pretrained model (such as DALL-E \cite{ramesh2021zero}) on a sequential text-to-image generation task, with the additional flexibility to copy from a prior input. To do so, we first `retro-fit' the model with additional layers to copy relevant output from the initial scene. Then, we introduce a self-attention block for generating story embeddings that provide global semantic context of the story during generation of each frame. The model is finetuned on the story continuation task, where these additional modules are trained from scratch. We name this approach \sdalle{} and also compare with a GAN-based model \sgan{} for story continuation. We also explore the parameter-efficient framework of prompt-tuning and introduce a prompt consisting of task-specific embeddings to coax the pretrained model into generating visualizations for the target domain. During training of this prompt-tuning version of the model, the pretrained weights are frozen and the new parameters are learned from scratch, which is time as well as memory-efficient.

Results show that our retro-fitting approach in \sdalle{} is useful for leveraging the latent pretrained knowledge of DALL-E for the story continuation task, and outperforms the GAN-based model on several metrics. Further, we find that the copying mechanism allows for improved generation in low-resource scenarios and of unseen characters during inference. In summary,
\begin{itemize}
    \item We introduce the task of story continuation, that is more closely aligned with real-world downstream applications for story visualization, and provide the community with a new story continuation dataset.
    \item We introduce \sdalle{}, an adaptation of pretrained transformers for story continuation, using retro-fitting. We also develop \sgan{} as a strong GAN baseline for comparison.
    \item We perform comparative experiments and ablations to show that finetuned \sdalle{} outperforms \sgan{} on three story continuation datasets along several metrics.
    \item Our analysis shows that the copying mechanism improves correlation of the generated images with the source image, leading to better continuity in the visual story and generation of low-resource as well as unseen characters.
\end{itemize}

\section{Related Work}

\paragraph{Text-to-Image Synthesis.}
Most work in text-to-image synthesis has focused on the development of increasingly sophisticated generative adversarial networks (GANs) \cite{goodfellow2014generative}. Recent works have leveraged multi-stage generation\cite{han2017stackgan}, attentional generative networks \cite{xu2018attngan}, dual learning \cite{qiao2019mirrorgan}, dynamic memory \cite{zhu2019dm, liang2019cpgan}, semantic disentaglement \cite{yin2019semantics}, explicit object modelling \cite{hinz2020semantic} and contrastive loss \cite{zhang2021cross, kang2020contragan} to further push performance on this task. DALL-E \cite{ramesh2021zero} is a large transformer language model that generates both text tokens and image tokens. VideoGPT \cite{yan2021videogpt} adapts the DALL-E architecture for conditional generation of videos from a first frame and trains it from scratch. In contrast, we adapt the pretrained DALL-E by \textit{retro-fitting} the pretrained weights with task-specific modules for conditional generation of a sequence of images from a first frame.

\paragraph{Story Visualization.}
\cite{li2019storygan} introduce the CLEVR-SV and PororoSV datasets which are based on the CLEVR \cite{johnson2017clevr} visual question answering dataset and Pororo video question answering dataset \cite{kim2017deepstory} respectively. \cite{maharana2021integrating} adapt the Flintstones text-to-video synthesis dataset \cite{gupta2018imagine} into FlintstonesSV. While these datasets have served as challenging benchmarks, they contain recurring characters throughout the dataset. Complex datasets, requiring story visualization models to generalize to a more diverse set of test cases are needed to better guide research in this domain. We introduce the story continuation task and propose a new dataset for the task.

Most story visualization models follow the framework introduced in StoryGAN\cite{li2019storygan}, which comprises a recurrent text encoder, an image generator, and image as well as story discriminators to train the GAN \cite{szHucs2022modular}. \cite{zeng2019pororogan} add textual alignment models and a path-based image discriminator, while \cite{LI2020102956} add dilated convolution and weighted activation degree to the discriminators. \cite{song2020character} add figure-background segmentation to the model in the form of generators and discriminators. \cite{maharana2021improving} and \cite{maharana2021integrating} use dual learning and structured inputs respectively to improve story visualization. We use their models as starting point and add modifications that leverage pretrained transformers for our proposed story continuation task.

\paragraph{Parameter-Efficient Training.}
Methods like adapter-tuning \cite{henderson2021compacter, mahabadi2021parameter, houlsby2019parameter, sung2022vl} and prompt-based tuning \cite{li2021prefix, lester2021power} add a small number of trainable parameters to the frozen weights of a pretrained model, which are then learned for the target task. Sparse updating  of parameters \cite{guo2021parameter, zaken2022bitfit} and low-rank decomposition matrices \cite{hu2021lora} also provide parameter-efficient methods for finetuning. \cite{mao2022unipelt, he2021towards} coaobine these approaches for a unified approach to finetuning pretrained models. \cite{borgeaud2022improving} `retro-fit' a pre-trained language model with cross-attention layers to retrieve relevant tokens at each timestep of word prediction in natural language generation. We use retro-fitting and prompt-tuning to adapt a pretrained image synthesis model to story continuation.

\section{Methods} \label{sec:modeling}
As discussed in Sec.~\ref{sec:intro}, story visualization has limited applicability in real-world settings because the task formulation does not allow models to generalize to new story elements. Hence, we propose the story continuation task and present our \sdalle{} and \sgan{} models for the task.

\subsection{Story Continuation}
Given a sequence of sentences $S=[s_1, s_2, ..., s_T]$ forming a narrative, story visualization is the task of generating a corresponding sequence of images $\hat{X} = [\hat{x}_1, \hat{x}_2, ..., \hat{x}_T]$, following \cite{li2019storygan}. $S$ contains a story, where the captions are temporally ordered and describe the same narrative. This task has many different potential applications such as facilitating the creation of comics or creating visualizations in an educational setting. However, due to the way that the story visualization task is formulated, current models are far from being applied to these settings. The models rely on the images seen in the training data, to generate new visualizations for input stories during the inference phase. Thus, they can only recreate the characters as already found in the training set. Additionally, the captions in story visualization datasets are focused on the narrative, which limits the amount of information that is provided to the model, including descriptions of characters or settings, background etc. Much of this is inferred by the model, leading to generations that might be drastically different than expected, and it is unrealistic to expect the models to generate completely new visual attributes without sufficient instructions in the caption. Story continuation addresses these issues by providing initial information about the story setting and characters.

In the story continuation task, the first image of the sequence $x_1$ is provided as additional input to the model. By including an initial ground truth scene as input, the model has access to the appearances of characters, the setting in which the story takes place, and more. When making subsequent scenes, the model then no longer needs to create all the visual features from scratch, but can instead copy from the initial frame. This first image addresses both the generalization issue and the limited information issue in current story visualization models. We refer to this first frame as \textit{source frame} and the remaining frames in the sequence $[x_2, ....., x_t]$ as \textit{target frames}.

\begin{figure*}[t]
    \centering
    \includegraphics[width=1.0\textwidth]{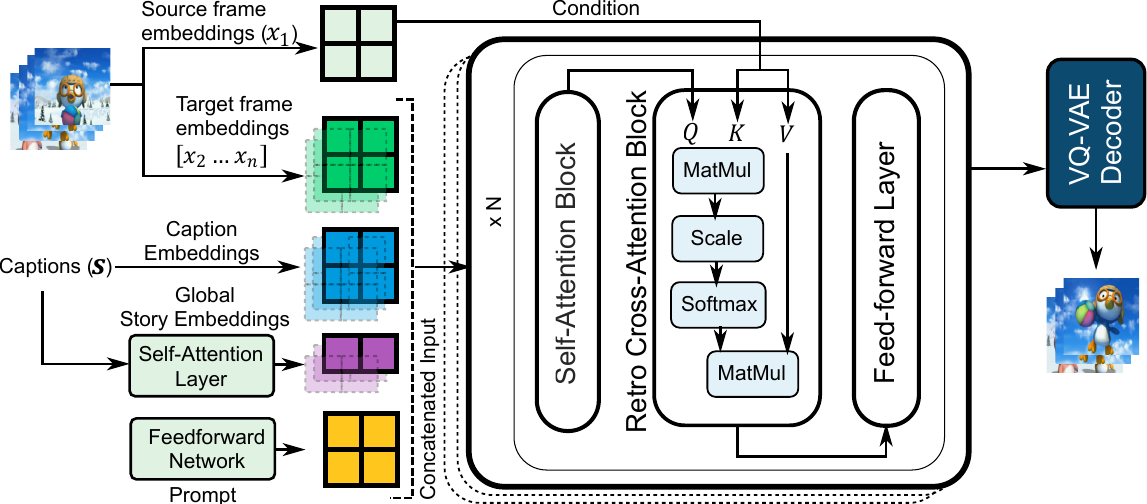}
    \caption{Illustration of our \sdalle{} architecture for the prompt-tuning setting. The frames are encoded using pretrained VQVAE and sent as inputs to the pretrained DALL-E. The inputs are prepended with input-agnostic prompt (in prompt-tuning setting only) and global story embeddings corresponding to each sample in the story continuation dataset. The output of \sdalle{} is decoded using VQ-VAE to generate the predicted image.
    \label{fig:model_storydalle}}
\end{figure*}

\subsection{\sdalle{}}
\label{sec:storydalle}

The DALL-E generative model is trained using a simple language-modeling objective on the sequence of discrete image tokens for the task of text-to-image synthesis \cite{ramesh2021zero}. With massive amounts of data, such models learn the implicit alignment between text tokens and image tokens, which can be leveraged for downstream tasks like story continuation. The two main aspects that differentiate the story continuation task from text-to-image synthesis are: (1) sequence of captions vs. single caption, and (2) source frame vs. no source frame. Hence, in order to convert the text-to-image synthesis model into a story continuation model, we add two task-specific modules to the native DALL-E architecture. First, we use a global story encoder to pool information from all captions and produce a story embedding, which provides global context of the story at each timestep. Next, we `retro-fit' the model with cross-attention layers in order to accept the source frame as additional input. We refer to our proposed model as \sdalle{} (see Figure~\ref{fig:model_storydalle}). All parameters of \sdalle{} are updated during the finetuning of the model. In the parameter-efficient version of \sdalle{}, we learn a sequence of embeddings for the story continuation task and provide it as a prompt to the model for task-specific instructions. During training, the pretrained model weights are frozen and these task-specific modules are trained from scratch.

\paragraph{Global Story Encoder.}
Most previous works in story visualization utilize recurrent encoders in the form of LSTM networks \cite{li2019storygan} or memory-augmented encoders \cite{maharana2021improving, maharana2021integrating}, to accept a sequence of captions as input. However, recurrent architectures are memory as well as time-intensive because of sequential processing. Hence, we propose to use a self-attention ($f_{self}$) based global story encoder, which takes the sentence embeddings for all captions as input and generates contextualized story embeddings for each time-step using parallel processing (see Figure~\ref{fig:model_storydalle}). Additionally, we initialize sinusoid positional embeddings ($S_{pos}$) to provide information about the position of the target frame within the story, and add those to the story embeddings: $S_{global} = f_{self}(S + S_{pos})$. These embeddings are prepended to the word embeddings for the caption at that timestep and sent as input to the generative model.

\paragraph{Retro-fitted Cross-Attention Blocks.}

Next, we want to `retro-fit' the DALL-E model with the ability to copy relevant elements from the source image, in order to promote generalizability to unseen visual attributes. This will allow the model to generate visual stories with completely new characters, as long as they are present in the source frame. Hence, we adapt the model to `condition' the generation of target frame on the source frame by adding a cross-attention block to each self-attention block of the native DALL-E architecture. The image embeddings of the source frame are used in the cross-attention layer as \textit{key} ($K$) and \textit{value} ($V$), while the output from the preceding self-attention layer is used as \textit{query} ($Q$). As shown in Figure~\ref{fig:model_storydalle}, the DALL-E self-attention block consists of the self-attention ($f_{self}^i$), feed-forward ($f_{dense}^{i}$) and normalization ($f_{norm}$) layers. Given an input $z_i$ to the $i$th self-attention block, the output $z^{i+1}$ is: $z^{i+1} = f_{norm}(f_{dense}^{i}(f_{self}^{i}(z_i)))$. In \sdalle{}, we insert a cross-attention layer such that the output $z^{i+1}$ is:
\begin{equation}
    z^{i+1} = f_{norm}(f_{dense}^{i}(f_{cross}^{i}(f_{self}^{i}(z^i), {c_{img}})))
\end{equation}
where $f_{cross}^{i}$ is the cross-attention layer in the $i$th transformer block and $c_{image}$ is the sequence of embedding representations for the conditioning image. The self-attention layers are constrained to perform causal masking for computing attention weights due to the nature of the image synthesis task. However, within the cross-attention layer, the input is free to attend over the entire source frame which eases the next token prediction task by augmenting the model with relevant information. The cross-attention layers are trained from scratch.\\

The \sdalle{} architecture can be fully fine-tuned to learn the weights of the above-mentioned task-specific modules, while updating the weights of the pretrained model as necessary, on the target task as well as dataset. However, \cite{borgeaud2022improving} show that freezing of pretrained weights during training of retro-fitted models can also lead to similar performance as models trained from scratch, with lesser training data. Further, it provides a parameter-efficient approach that can be trained/deployed with a smaller amount of computational resources. Hence, we additionally explore prompt-tuning \cite{li2021prefix} of the \sdalle{} model.

\paragraph{Prompt.} Prompt-tuning is an alternative \cite{li2021prefix} to full model fine-tuning where the pretrained model weights are frozen and instead, a small sequence of task-specific vectors is optimized for the downstream task. We initialize a parameterization network $MLP(.)$, which takes a matrix of trainable parameters $P_\theta^{'}$ of dimensions $P_{idx}$ and $dim(h^i)$ as input and generates the prompt $P_\theta$. These trainable matrices are randomly initialized and trained from scratch on the downstream task and dataset. $P_\theta$ is appended to the word embeddings of input caption, along with the global story embeddings. Together, these additional embedding vectors act as `virtual tokens' of a task-specific prompt, and are attended to by each of the caption as image tokens. Formally, the input $h^{i}$ to the $i$th self-attention layer in the auto-regressive transformer is organized as follows:
\begin{equation}
    h^i = 
    \begin{cases}
    P_{\theta}[j, :] & \text{if}\; j\in [0, P_{idx})\\
    S_{global} & \text{if}\; j== P_{idx} \\
    f^{i}(z_j, h_{<j}) & \text{otherwise}
    \end{cases}
\end{equation}
where $f^{i}(.)$ is the $i$th transformer block in \sdalle{}.

With the aforementioned additions, we convert the pretrained DALL-E into \sdalle{} model for the story continuation task. A pretrained VQVAE encoder \cite{van2016conditional} is used to transform RGB images into small 2D grids of image tokens, which are flattened and concatenated with the modified inputs in \sdalle{} (see Appendix for details). Finally, \sdalle{} is trained to model the joint distribution over the tokens of text $s$ and image $x$: $p(x) = \prod_{j=1}^{d} p(x_{j}|x_{<i};s)$. New parameters as well as pretrained weights are optimized in full-model finetuning whereas only the parameters of the prompt, story encoder and cross-attention layers are optimized during prompt-tuning.

\subsection{\sgan{}}
Generative Adversarial Networks (GANs) have enjoyed steady progress at many image generation tasks such as style transfer \cite{karras2019style}, conditional image generation \cite{xu2018attngan}, image-to-image translation \cite{isola2017image} over the last decade. Unlike transformers, they do not need to be pretrained on massive datasets, and can be trained for narrow domains with smaller datasets, which makes it an appealing method. Several recent works in story visualization have demonstrated the effectiveness of GANs for this task \cite{li2019storygan, maharana2021improving, song2020CPCSV}. Hence, we also develop a GAN-based model, \sgan{}, for the story continuation task and compare its performance to that of \sdalle{} on the proposed datasets (see Appendix for figure and details). \sgan{} follows the general framework of the StoryGAN model \cite{li2019storygan} i.e., it is composed of a recurrent text encoder, an image generation module, and two discriminators - image and story discriminator. We modify this framework to accept the source frame as input for the story continuation task, and use it for improving the generation of target frames. Our \sgan{} model is implemented as follows:

\paragraph{Pre-trained Language Model Encoder.}
We use a pretrained language model (such as RoBERTa \cite{liu2019roberta} or CLIP text encoder \cite{radford2021learning}) as the caption encoder. These models are pretrained on large unimodal or multimodal datasets of language, which is of great utility for understanding the semantic concepts present in input captions. To ensure that the model has access to all captions, we append the captions together and use a special token to denote which caption is currently being generated.

\paragraph{Contextual Attention.}
The story representation from the encoder is combined with the image embeddings of the first frame of the image sequence using contextual attention \cite{yu2018generative} between the two inputs. The resulting representation is fed through a generator module which recurrently processes each caption, and produces a corresponding image.

\paragraph{Discriminators.}
The story discriminator takes all of the generated images and uses 3D convolution to create a single representation and then makes a prediction as to whether the generated story is real or fake. The image discriminator performs the same function but only focuses on individual images. The KL-Divergence loss enforces gaussian distribution on the latent representations learnt by GAN. Finally, the model is trained end-to-end using the objective function: $\min_{\theta_{G}} \max_{\theta_{I},\theta_{S}} \> \mathcal{L}_{KL} + \mathcal{L}_{img} + \mathcal{L}_{story}$, where $\theta_{G}$, $\theta_{I}$ and $\theta_{S}$ denote the parameters of the text encoder $+$ image generator, and image and story discriminators respectively. During inference, the trained weights $\theta_{G}$ are used to generate a visual story for a given input of captions.

\section{Datasets} \label{sec:datasets}

Since story continuation is a reframing of the story visualization tasks, existing story visualization datasets can be adapted for story continuation by assigning the first frame in the sequence as source frame and the rest as target frames. However, such existing story visualization datasets like PororoSV \cite{li2019storygan} and FlintstonesSV \cite{gupta2018imagine} are also homogeneous datasets with recurring characters i.e., the characters used during evaluation already appear in the training set. It is not possible to evaluate the generalization capacity of story continuation models using these datasets. Hence, we propose a new dataset in this paper.

\begin{figure*}[t]
    \centering
    \includegraphics[width=0.65\textwidth]{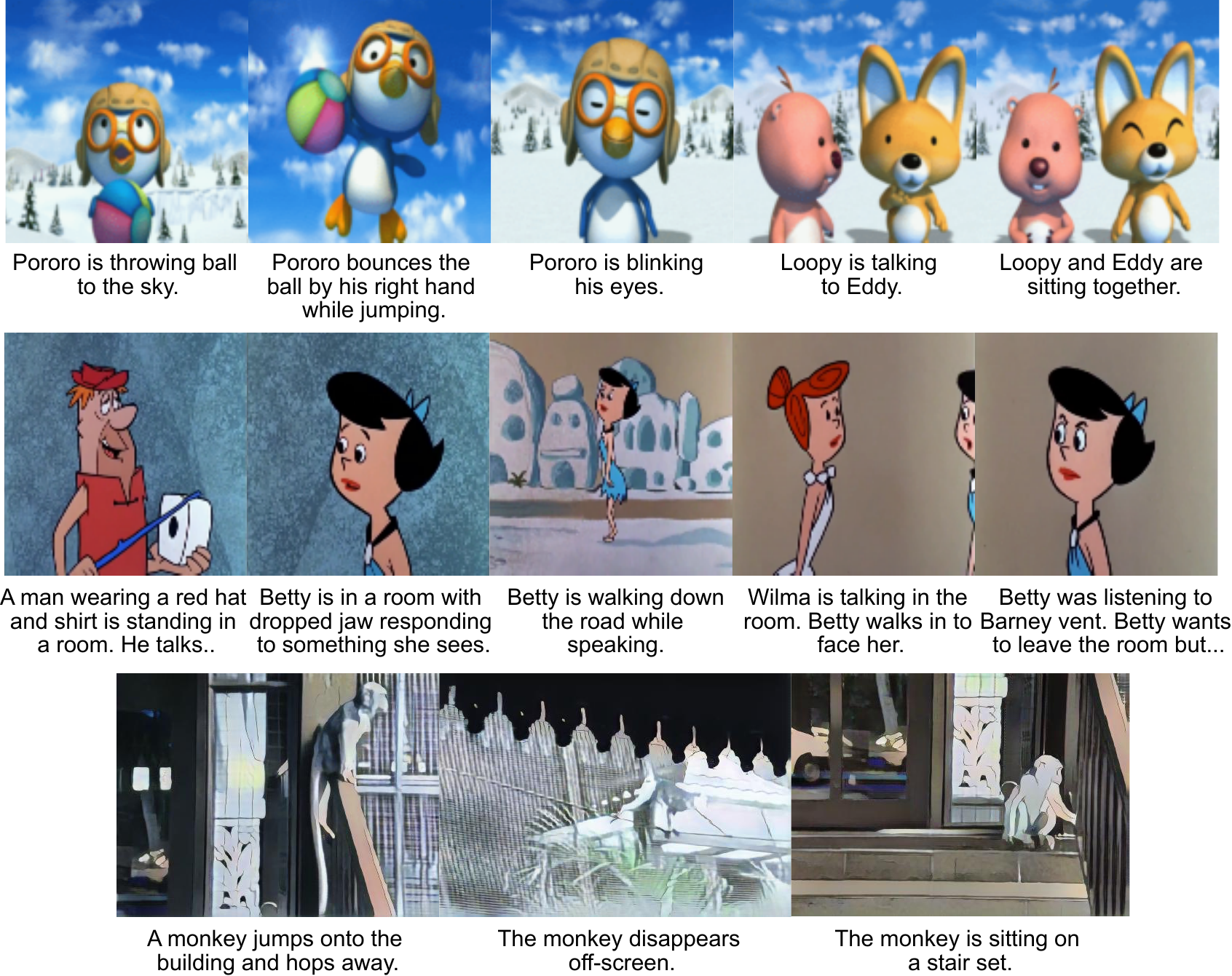}
    \caption{Examples from the PororoSV (top), FlintstonesSV (middle) and DiDeMoSV (bottom) datasets. In the story continuation setting, the first frame is used as input to the generative model.}
    \label{fig:example}
\end{figure*}
% \end{wrapfigure}

\paragraph{DiDeMoSV.}
DiDeMo \cite{hendricks17iccv} is a video captioning dataset containing ~10,000 short clips with more than 40,000 text descriptions temporally localized with the videos. Each of the clips was randomly sampled from the YFCC100M \cite{thomee2016yfcc100m} dataset which is based upon Flickr. This results in videos that cover a large breadth of real-world scenarios, containing many different settings, actions, entities, and more. The dataset contains 11550/2707/3378 samples in training, validation and test respectively, with each sample containing three consecutive frames. This dataset challenges story continuation models to generate diverse inputs, covering many more story elements, in contrast to existing story visualization datasets. In order to do this, models must maximize their usage of the initial scene input and need to incorporate additional general visual knowledge, whether this is done through transfer learning or additional data.

We also use the existing PororoSV \cite{li2019storygan} and FlintstonesSV datasets \cite{gupta2018imagine}, containing 10191/2334/2208 and 20132/2071/2309 samples respectively, to evaluate our story continuation models. Each sample contains 5 consecutive frames. There are 9 and 7 main characters in PororoSV and FlintstonesSV respectively, that appear throughout the dataset. For story continuation, we use the first frame as source frame and the rest of the four frames in the sequence as target frames. Evaluation is only performed on the generation of target frames. See Figure~\ref{fig:example} for examples from the three story continuation datasets.

\section{Experiments}
We use the pretrained weights from popular open-source minDALL-E (1.3B parameters) which is trained on 14 million text-image pairs from the CC3M \cite{sharma2018conceptual} and CC12M \cite{changpinyo2021conceptual} datasets, to initialize our models.\footnote{\url{https://github.com/kakaobrain/minDALL-E}}
minDALL-E uses the pretrained VQGAN-VAE \cite{esser2021taming} for discretizing image inputs. We experiment with pretrained CLIP \cite{radford2021learning} (38M parameters) and distilBERT \cite{sanh2019distil} (110M parameters) text encoders for the \sgan{} models. The \sdalle{} models are trained for 5 epochs with learning rates of 1e-04 (AdamW, Cosine Scheduler) and 5e-04 (AdamW, Linear Decay Scheduler) for full-model fine-tuning and prompt-tuning setups respectively. Checkpoints are saved at the end of every epoch. The \sgan{} models are trained for 120 epochs with learning rates 1e-04 and 1e-05 for the generator and discriminators respectively. Checkpoints are saved every 10 epochs. These models are trained on single A6000 GPUs. 

We use the FID score for saving the best checkpoints in our experiments. The FID score calculates the difference between the ground truth and generated images by computing the distance between two feature vectors. Following \cite{li2019storygan} and \cite{maharana2021improving}, we also compute the character classification scores (F1 Score and Frame Acc.) for the PororoSV and FlintstonesSV datasets. See Appendix for details.

\begin{table*}[t]
% \small
\centering
\caption{\label{tab:test_fid}Results on the test sets of PororoSV, FlintstonesSV and DiDeMoSV (DSV) datasets from various models. Scores are based on FID (lower is better), character classification F1, and frame accuracy (F-Acc.; higher is better) evaluations.}
\scalebox{1.0}{
\begin{tabular}{|c|c|c|c|c|c|c|c|}
\hline
\multirow{2}{*}{\textbf{Model}} & \multicolumn{3}{c|}{\textbf{ PororoSV }} & \multicolumn{3}{c|}{\textbf{ FlintstonesSV }} & \textbf{DSV}\\
\cline{2-8}
 & FID $\downarrow$&{Char-F1$\uparrow$}& F-Acc$\uparrow$& FID $\downarrow$&{Char-F1$\uparrow$}& F-Acc$\uparrow$ & FID$\downarrow$\\
\hline
\sgan{} (BERT) & 72.98 & \textbf{43.22} & 17.09 & 91.37 & 70.45 & 55.78 & 91.43 \\
\sgan{} (CLIP) & 74.63 & 39.68 & 16.57 & 90.29 & 72.80 & \textbf{58.39} & 92.64\\
\sdalle{} (prompt) & 61.23 & 29.68 & 11.65 & 53.71 & 42.48 & 32.54 & 64.58 \\
\sdalle{} (finetuning) & \textbf{25.90} & 36.97 & \textbf{17.26} & \textbf{26.49} & \textbf{73.43} & 55.19 & \textbf{32.92} \\
\hline
\end{tabular}
}
\end{table*}

\section{Results}
\label{sec:results}

\paragraph{Main Quantitative Results.}
Table~\ref{tab:test_fid} contains the FID, character classification F1 score and frame accuracy results on the test sets of PororoSV and FlintstonesSV datasets using various models in our experiments. We train two variations of the \sgan{} model with the distilBERT and CLIP text encoders. Our model \sdalle{} is trained under two settings, one where the pretrained weights are frozen during training and the other where the pretrained weights are also finetuned on the target dataset. In practice, we find it necessary to finetune the pretrained text and image embeddings within the transformers, which are pretrained on real-world images, in the prompt tuning setting in order to adapt them to different domains such as cartoons. This results in nearly 30\% trainable parameters during prompt-tuning, as compared to full-model finetuning. With fully finetuned \sdalle{}, we see drastic improvements in FID score for the PororoSV and FlinstonesSV datasets, over the \sgan{} model, demonstrating the superior visual quality of the generated visual stories. The character classification scores remain the same for FlintstonesSV and drop by 6\% and 14\% for PororoSV with use of finetuned and prompt-tuned \sdalle{} respectively. GAN-based models like \sgan{} are able to recreate distinct and finer details of a character which leads to higher accuracy scores using a classification model, such as the Inception-v3 used in our experiments \cite{maharana2021improving}. With prompt-tuning, we observe that \sdalle{} models manage to capture the background elements of the scene but fail to properly recreate the characters in the frame. The frame accuracy score, which is based on exact match overlap of multiple characters in the predicted scene with those in ground truth, remains low for all models, suggesting that both methods struggle to compose multiple roles in a single image \cite{cho2022dall}. 

For the more challenging DiDeMoSV dataset, the fully finetuned \sdalle{} model outperforms the GAN models by a wide margin in terms of FID score. It should be noted here that PororoSV and FlintstonesSV have a finite set of recurring animated characters throughout the dataset, whereas DiDeMoSV is derived from a multitude of real-world scenarios with no overlap in characters between training and evaluation sets. While the addition of a source frame makes it easier for the model to replicate it in the target frames, the generation is significantly more difficult due to the diversity in evaluation samples. However, since the DiDeMoSV dataset contains images from the real-world domain, the pretrained knowledge of \sdalle{} derived from Conceptual Captions is useful for generating relevant and coherent images for the dataset, while \sgan{} largely fails to do so.

\paragraph{Ablations.}

\begin{table*}[t]
% \small
\centering
\caption{\label{tab:val_ablation}Ablation results of finetuned StoryDALL-E on validation sets of PororoSV, FlintstonesSV and DiDeMoSV (DSV) datasets. Scores are based on FID (lower is better), character classification F1 and frame accuracy (F-Acc.; higher is better) evaluations.}
\scalebox{0.85}{
\begin{tabular}{|l|c|c|c|c|c|c|c|}
\hline
\multirow{2}{*}{\textbf{         Model}} & \multicolumn{3}{c|}{\textbf{ PororoSV }} & \multicolumn{3}{c|}{\textbf{ FlintstonesSV }} & \textbf{DSV}\\
\cline{2-8}
 & FID $\downarrow$&{Char-F1$\uparrow$}& F-Acc$\uparrow$& FID $\downarrow$&{Char-F1$\uparrow$}& F-Acc$\uparrow$ & FID$\downarrow$\\
\hline
\sdalle{} & 21.64 & 40.28 & 20.94 & 28.37 & 74.28 & 52.35 & 41.58 \\
- Cross-Attention & 30.45 & 39.32 & 34.65 & 35.04 & 73.94 & 53.28 & 55.89\\
- Story Embeddings & 23.27 & 40.25 & 18.16 & 29.21 & 72.18 & 52.72 & 42.34 \\
- Story Embeddings \& Cross-Attention & 31.68 & 35.29 & 16.73 & 36.28 & 72.44 & 51.32 & 58.14 \\
\hline
\end{tabular}
}
\end{table*}

Table~\ref{tab:val_ablation} contains results from ablation experiments on finetuned StoryDALL-E on the validation sets of the three story continuation datasets. The primary modifications we make to DALL-E in order to adapt it into \sdalle{}, are the cross-attention layers and global story embeddings. We perform minus-one experiments on StoryDALL-E by removing each of these components and observing the effect on FID results on validation sets. First, we remove the cross-attention layers from StoryDALL-E, which reverts the model to the story visualization setting where the model no longer receives the first image as input, and is evaluated on the generation of the rest of the frames in the visual story. With this ablation, we see a large increase in FID scores across all datasets. Without a source image to guide the generated output, the quality of illustration drops rapidly, especially for the new DiDeMo dataset. The removal of global story embeddings results in a text-to-image synthesis setting with the first frame as additional input. In this scenario, we see smaller drops in FID, indicating that the global context is not as important as the ability to copy from an initial image. In the third row, we remove both, cross-attention layers and story embeddings, which relegates the setting to a text-to-image synthesis task, and observe a large increase in FID scores across all datasets.

\begin{table*}[t]
\centering
\caption{Results from human evaluation (Win\% / Lose\% / Tie\%). Win\% = \% times stories from \sdalle{} was preferred over \sgan{}, Lose\% for vice-versa. Tie\% represents remaining samples.}
\scalebox{1.0}{
\begin{tabular}{|c|c|c|c|}
\hline
\textbf{Dataset} & \textbf{~Visual Quality~} & \textbf{~Relevance~} & \textbf{~Consistency~}  \\ 
\hline
    PororoSV & 94/0/6  & 44/28/28 & 56/26/18 \\
    FlintstonesSV & 90/2/8 & 32/38/30 & 42/32/26  \\
  DiDeMoSV & 64/0/36 & 38/0/62 & 32/48/20  \\
\hline
\end{tabular}
}
\label{tab:human_results}
\end{table*}

\subsection{Human Evaluation}
We additionally conduct human evaluation on our models' outputs hoping to better capture the overall quality of the generated stories. We have a human annotator compare generated visual stories from our \sdalle{} (finetuning) and \sgan{} (BERT) models. They are provided with predictions from each dataset and the corresponding ground truth captions and asked to pick the better prediction (or tie) in terms of visual quality, consistency, and relevance \cite{li2019storygan}. Results are presented in Table~\ref{tab:human_results}. The \sdalle{} model outperforms \sgan{} model in terms of visual quality and relevance, achieving higher \% of wins in each of the three datasets (except relevance in FlintstonesSV). These results follow from the fact that \sdalle{} uses the VQGAN-VAE \cite{esser2021taming} which is designed for reconstructing higher resolution images. Moreover, it has access to large pretraining data, which improves alignment between semantic concepts in captions and regions in images. We see wins in terms of consistency for PororoSV and DiDeMoSV predictions from \sdalle{} models. But, the absolute numbers for consistency and relevance show that there is still room for improvement.

\begin{figure*}[t]
    \centering
    \includegraphics[width=0.96\textwidth]{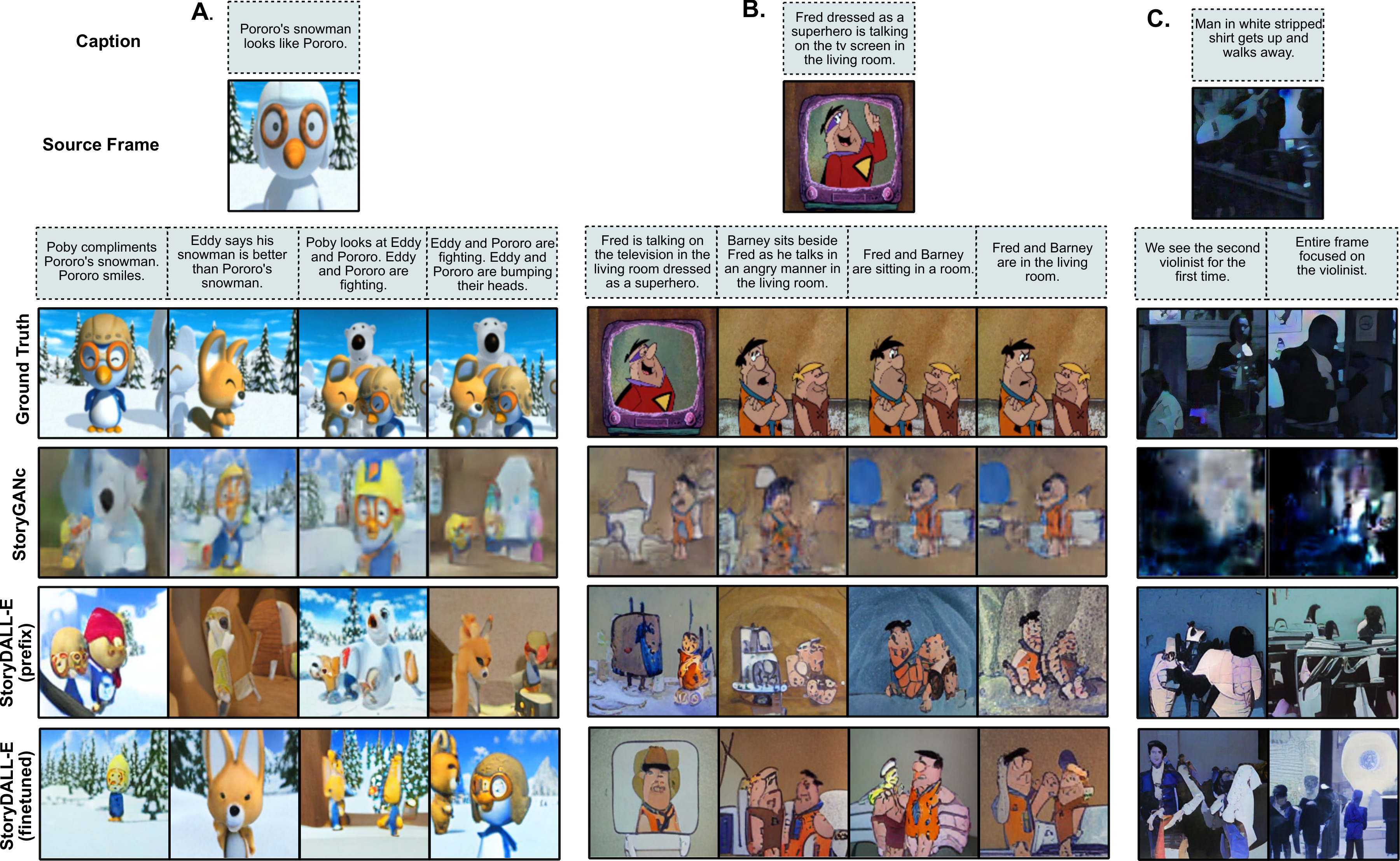}
    \caption{Examples of predictions for (A) PororoSV (B) FlintstonesSV and (C) DiDeMoSV story continuation datasets from finetuned \sdalle{} and \sgan{} models. Source frame refers to the initial frame provided as additional input to the model.}
    \label{fig:predictions}
\end{figure*}

\section{Analysis}
In this section, we perform experiments to analyze aspects of the \sdalle{} model and the story continuation task. First, we perform qualitative analyses of the predictions from \sdalle{}. Next, we quantify the effect of the retro-fitted cross-attention layers and visualize the attention heads. See Appendix for an analysis of the diverse semantic content in the DiDeMoSV dataset. 

\subsection{Qualitative Analysis}
\label{sec:qual_analysis}

Figure~\ref{fig:predictions} contains sampled outputs from both of our models for the three story continuation datasets. In each of these examples, \sdalle{} generates higher quality images than \sgan{}. The difference is especially stark for PororoSV and FlintstonesSV datasets since \sdalle{} is exposed to the characters during training and has additional guidance from source frame during inference. In the case of DiDeMoSV, the generations from \sgan{} are largely incomprehensible, which could be attributed to the unseen semantic concepts such as `violinist' which did not appear in the training set. In contrast, \sdalle{} is exposed to various real-world concepts during pretraining, which can be leveraged during generation. For instance, the pretrained knowledge, as well as the copying mechanism, help the \sdalle{} model comprehend `television' and generate an image for `Fred is talking in the television' (see Figure~\ref{fig:predictions}(b)). However, the overall quality of the images from \sdalle{} also does not approach human produced images. As discussed in Sec.~\ref{sec:results}, it is especially true for frames containing multiple characters. This suggests that while current models are able to attempt the task, there is still much work to be done before consistent and coherent images are commonly produced by the models.

We also examine the ability of \sdalle{} to recreate scarce characters from the training set (see Figure~\ref{fig:cross_attn_viz}(a)) and generate unseen characters (see Figure~\ref{fig:cross_attn_viz}(b)), when guided by the copying mechanism via cross-attention layers. We find that the copying mechanism allows for better generation of shape and form for less-frequent characters in PororoSV. Similarly, we identified non-recurring characters in the FlintstonesSV dataset and observed the corresponding generated images, when \sdalle{} has access to a previous frame where they appear. \sdalle{} succeeds at partially copying visual aspects of the characters, such as the purple skirt (top) and blue uniform (bottom).

\subsection{Retro-fitted Cross-Attention}
\begin{figure*}[t]
    \centering
    \includegraphics[width=1.0\textwidth]{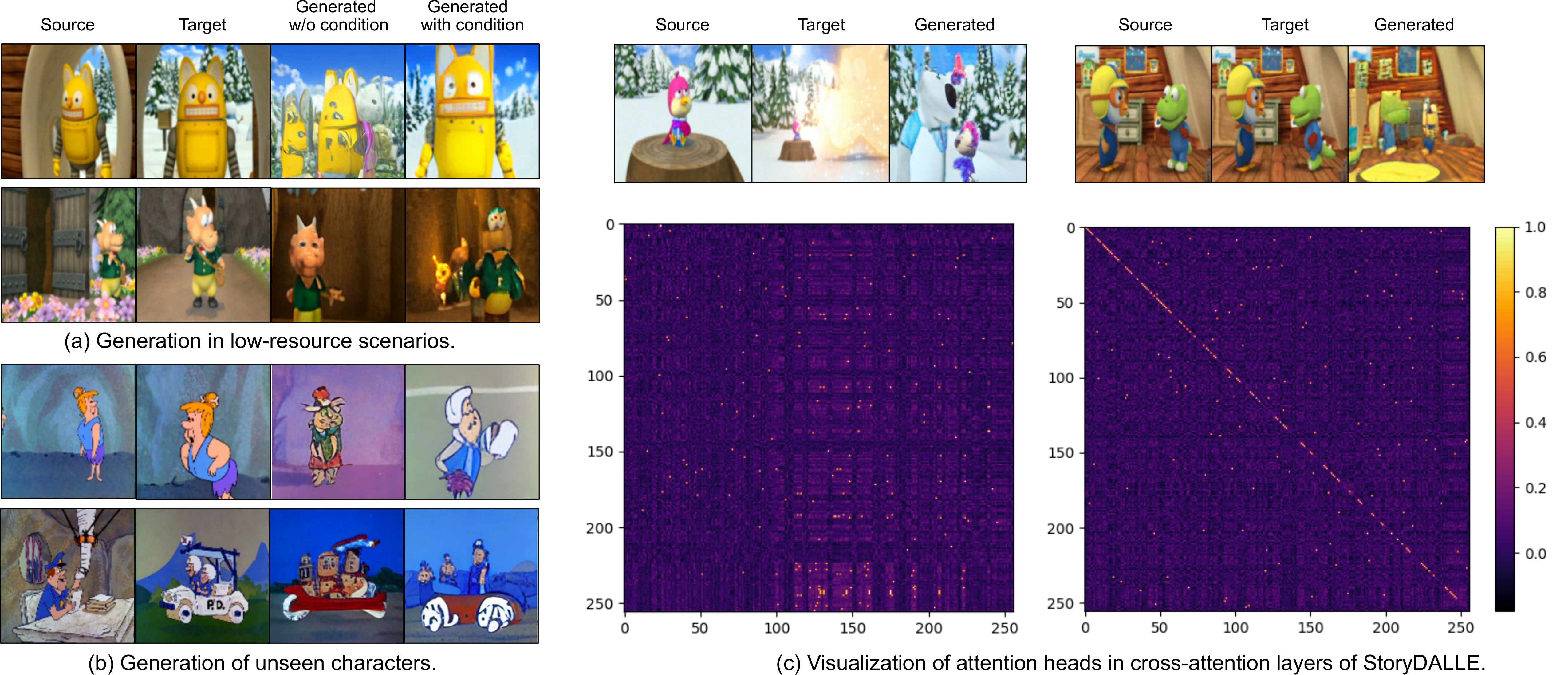}
            \caption{Examples of generation from \sdalle{} in (a) low-resource scenarios and (b) of unseen characters. (c) Plots of attention scores computed in retro cross-attention layers for examples of source frames (x-axis) and target frames (y-axis).
    \label{fig:cross_attn_viz}}
\end{figure*}

We examine the attention scores computed in the retro cross-attention layer and present examples in Figure~\ref{fig:cross_attn_viz}(c). The cross-attention layers in \sdalle{} receive vector representations for the source image and compute the cross-attention output using the source frame as key/value and the target frame as query. In the first example (left), the target frame is copying visual attributes of the pink bird with the most emphasis, as be seen from the higher attention scores for the image tokens roughly in the center of the source frame. For the second example (right), the source frame and target frames are nearly similar; the attention scores are highest in the diagonal of the plot. The resulting images in both samples contain many visual attributes already found in the source image, demonstrating that the cross-attention layer is effective at enabling conditional image generation. See Appendix for correlation scores between source image and frames generated with or without condition using \sdalle{}.

\section{New Results \& Demo with DALL-E Mega}
Following our approach of using pretrained text-to-image synthesis models for story continuation, we repeat our experiments with the recently released DALL-E Mega for the final version of the paper.\footnote{\url{https://github.com/kuprel/min-dalle}} DALL-E Mega is pretrained on 15 million images from the Conceptual Captions dataset \cite{sharma2018conceptual} and follows an encoder-decoder architecture, as opposed to the decoder-only architecture used in minDALL-E. It relies on the pretrained BART encoder \cite{lewis2020bart} for encoding the input captions as well as an improved VQGAN-VAE for discretized encoding of images. In order to adapt DALL-E Mega for story continuation, we retro-fit the decoder in the pretrained model with cross-attention layers and a global story encoder as outlined in Sec.~\ref{sec:modeling}. These additional cross-attention layers facilitate copying from a source image, and the story encoder enables generation of a sequence of frames for the story continuation task. We refer to the \sdalle{} model based on the pretrained DALL-E Mega as the \sdallem{} model in this paper. In the fully-finetuned version of the \sdallem{} model, we finetune the encoder as well as the decoder on story continuation datasets. Results are shown in Table~\ref{tab:mega_dalle}. We observe up to 3\% improvement in FID scores over \sdalle{} across all datasets. Smaller improvements are observed for character classification scores with the use of DALL-E Mega. Examples are shown in Figure~\ref{fig:predictions_mega}.

\begin{figure*}[t]
    \centering
    \includegraphics[width=0.96\textwidth]{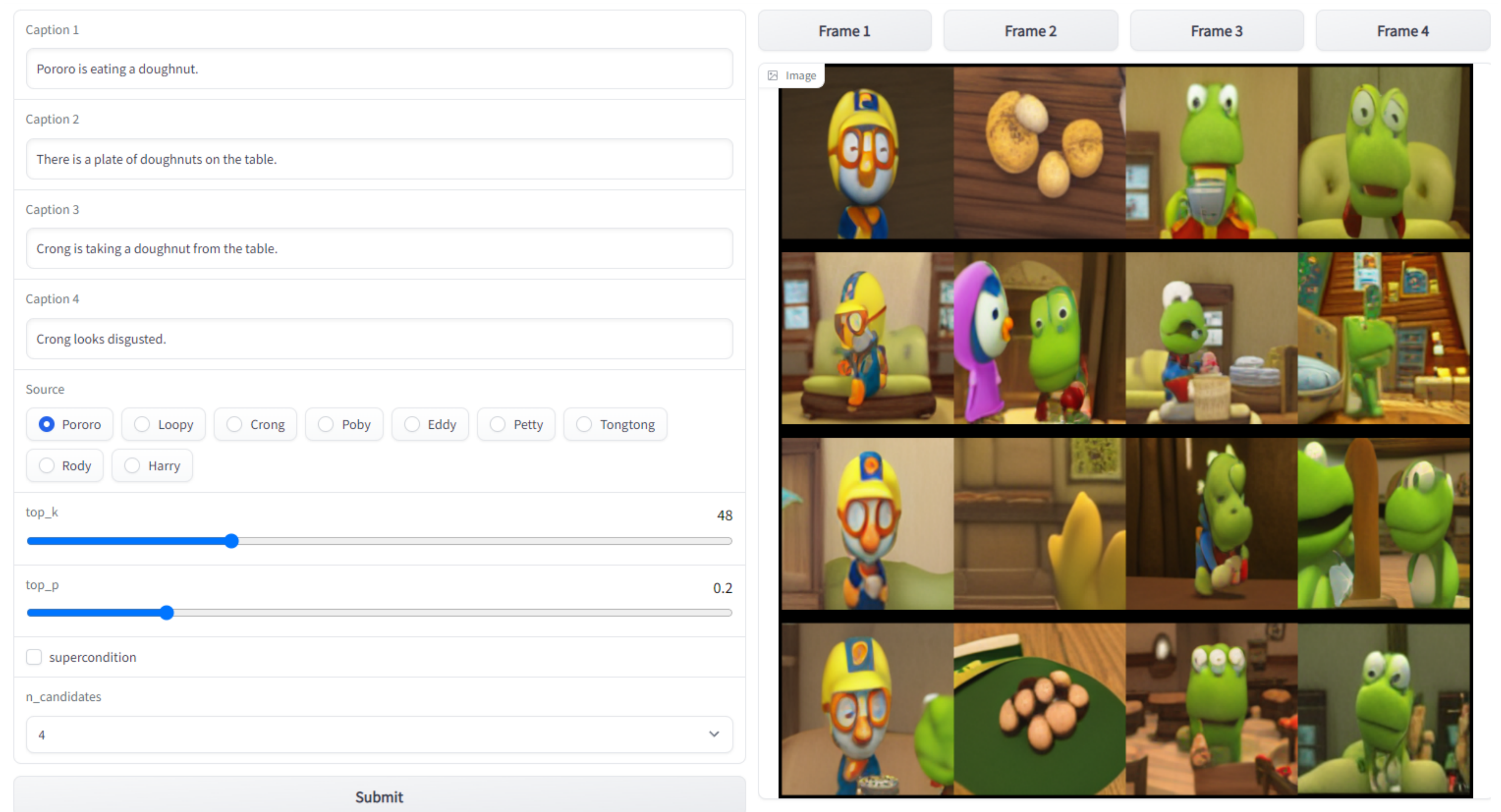}
    \caption{A snapshot of the openly-available in-browser demo for \sdallem{} trained on the Pororo dataset. The right panel displays the images generated by the model for the captions entered by the user in the left panel.}
    \label{fig:demo_mega}
\end{figure*}

We make the \sdallem{} model trained on the Pororo dataset available for testing through an openly accessible and easy-to-use in-browser demo system (see Figure~\ref{fig:demo_mega}).\footnote{See Model Card \cite{mitchell2019model} \& Demo at \url{https://github.com/adymaharana/storydalle}.} From Figures~\ref{fig:predictions_mega},~\ref{fig:demo_mega} and demo examples, we find that the model performs well at visualizing stories with up to three characters across all frames and struggles at generating coherent visuals for more than three characters, which is also in line with our findings in Sec.~\ref{sec:qual_analysis}. The model copies visual elements from the source image and copies to each of the generated frames in the story, hence maintaining a continuous flow in narration by virtue of conditioning on an initial scene. \sdallem{} performs best at generating overtly visual actions such as `making cookies', `walking', `reading a book'. Further, it is capable of generating semantic concepts that do not appear in the story continuation dataset, such as `doughnut' and `lion', by leveraging the pretrained knowledge of DALL-E Mega when possible. Most of the scenes in the Pororo dataset occur within the setting of a snowy village with wooden houses surrounded by trees and snow. Hence, the model usually generates scenes with similar visual elements.

\begin{table*}[t]
\small
\centering
\caption{\label{tab:mega_dalle}Results on the test sets of PororoSV, FlintstonesSV and DiDeMoSV (DSV) datasets from \sdallem{}. Scores are based on FID (lower is better), character classification F1 and frame accuracy (F-Acc.; higher is better) evaluations.}
\scalebox{0.9}{
\begin{tabular}{|c|c|c|c|c|c|c|c|}
\hline
\multirow{2}{*}{\textbf{Model}} & \multicolumn{3}{c|}{\textbf{ PororoSV }} & \multicolumn{3}{c|}{\textbf{ FlintstonesSV }} & \textbf{DSV}\\
\cline{2-8}
 & FID $\downarrow$&{Char-F1$\uparrow$}& F-Acc$\uparrow$& FID $\downarrow$&{Char-F1$\uparrow$}& F-Acc$\uparrow$ & FID$\downarrow$\\
\hline
\sdalle{} (finetuning) & 25.90 & 38.48 & 17.26 & 26.49 & 73.43 & \textbf{55.19} & 32.92 \\
\sdallem{} (finetuning) & \textbf{23.48} & \textbf{39.91} & \textbf{18.01} & \textbf{23.58} & \textbf{74.26} & 54.68 & \textbf{31.64} \\
\hline
\end{tabular}
}
\end{table*}

\section{Conclusion}
We introduce a new task called story continuation in order to make the story visualization task more conducive for real-world use cases. We present a new dataset DiDeMoSV, in addition to reformatting two existing story visualization datasets for story continuation. Our model \sdalle{}, based on a retro-fitting approach for adapting pretrained transformer-based text-to-image synthesis models, outperforms GAN-based models on the story continuation datasets. We also added new, improved results and a demo system using the more recent, larger DALL-E Mega model. We hope that the dataset and models motivate future work in story continuation and that our work encourages the exploration of text-to-image synthesis models for more complex image synthesis tasks.

\paragraph{Acknowledgement.} We thank the reviewers for their useful feedback. This work was supported by ARO Award W911NF2110220, DARPA KAIROS Grant FA8750-19-2-1004, NSF-AI Engage Institute DRL-211263. The views, opinions, and/or findings contained in this article are those of the authors, not the funding agency.

\bibliographystyle{splncs04}
\bibliography{egbib}

\begin{thebibliography}{10}
\providecommand{\url}[1]{\texttt{#1}}
\providecommand{\urlprefix}{URL }
\providecommand{\doi}[1]{https://doi.org/#1}

\bibitem{borgeaud2022improving}
Borgeaud, S., Mensch, A., Hoffmann, J., Cai, T., Rutherford, E., Millican, K.,
  Van Den~Driessche, G.B., Lespiau, J.B., Damoc, B., Clark, A., et~al.:
  Improving language models by retrieving from trillions of tokens. In:
  International Conference on Machine Learning. pp. 2206--2240. PMLR (2022)

\bibitem{cer2018universal}
Cer, D., Yang, Y., Kong, S.y., Hua, N., Limtiaco, N., John, R.S., Constant, N.,
  Guajardo-Cespedes, M., Yuan, S., Tar, C., et~al.: Universal sentence encoder.
  arXiv preprint arXiv:1803.11175  (2018)

\bibitem{changpinyo2021conceptual}
Changpinyo, S., Sharma, P., Ding, N., Soricut, R.: Conceptual 12m: Pushing
  web-scale image-text pre-training to recognize long-tail visual concepts. In:
  Proceedings of the IEEE/CVF Conference on Computer Vision and Pattern
  Recognition. pp. 3558--3568 (2021)

\bibitem{chen2018cartoongan}
Chen, Y., Lai, Y.K., Liu, Y.J.: Cartoongan: Generative adversarial networks for
  photo cartoonization. In: Proceedings of the IEEE conference on computer
  vision and pattern recognition. pp. 9465--9474 (2018)

\bibitem{cho2022dall}
Cho, J., Zala, A., Bansal, M.: Dall-eval: Probing the reasoning skills and
  social biases of text-to-image generative transformers. arXiv preprint
  arXiv:2202.04053  (2022)

\bibitem{esser2021taming}
Esser, P., Rombach, R., Ommer, B.: Taming transformers for high-resolution
  image synthesis. In: Proceedings of the IEEE/CVF Conference on Computer
  Vision and Pattern Recognition. pp. 12873--12883 (2021)

\bibitem{frans2021clipdraw}
Frans, K., Soros, L., Witkowski, O.: Clipdraw: Exploring text-to-drawing
  synthesis through language-image encoders. arXiv preprint arXiv:2106.14843
  (2021)

\bibitem{goodfellow2014generative}
Goodfellow, I.J., Pouget-Abadie, J., Mirza, M., Xu, B., Warde-Farley, D.,
  Ozair, S., Courville, A.C., Bengio, Y.: Generative adversarial nets. In:
  NeurIPS (2014)

\bibitem{guo2021parameter}
Guo, D., Rush, A.M., Kim, Y.: Parameter-efficient transfer learning with diff
  pruning. In: Proceedings of the 59th Annual Meeting of the Association for
  Computational Linguistics and the 11th International Joint Conference on
  Natural Language Processing (Volume 1: Long Papers). pp. 4884--4896 (2021)

\bibitem{gupta2018imagine}
Gupta, T., Schwenk, D., Farhadi, A., Hoiem, D., Kembhavi, A.: Imagine this!
  scripts to compositions to videos. In: Proceedings of the European Conference
  on Computer Vision (ECCV). pp. 598--613 (2018)

\bibitem{he2021towards}
He, J., Zhou, C., Ma, X., Berg-Kirkpatrick, T., Neubig, G.: Towards a unified
  view of parameter-efficient transfer learning. In: International Conference
  on Learning Representations (2021)

\bibitem{henderson2021compacter}
Henderson, J., Ruder, S., et~al.: Compacter: Efficient low-rank hypercomplex
  adapter layers. In: Advances in Neural Information Processing Systems (2021)

\bibitem{hendricks17iccv}
Hendricks, L.A., Wang, O., Shechtman, E., Sivic, J., Darrell, T., Russell, B.:
  Localizing moments in video with natural language. In: Proceedings of the
  IEEE International Conference on Computer Vision (ICCV) (2017)

\bibitem{hinz2020semantic}
Hinz, T., Heinrich, S., Wermter, S.: Semantic object accuracy for generative
  text-to-image synthesis. IEEE transactions on pattern analysis and machine
  intelligence  (2020)

\bibitem{houlsby2019parameter}
Houlsby, N., Giurgiu, A., Jastrzebski, S., Morrone, B., De~Laroussilhe, Q.,
  Gesmundo, A., Attariyan, M., Gelly, S.: Parameter-efficient transfer learning
  for nlp. In: International Conference on Machine Learning. pp. 2790--2799.
  PMLR (2019)

\bibitem{hu2021lora}
Hu, E.J., Wallis, P., Allen-Zhu, Z., Li, Y., Wang, S., Wang, L., Chen, W.,
  et~al.: Lora: Low-rank adaptation of large language models. In: International
  Conference on Learning Representations (2021)

\bibitem{isola2017image}
Isola, P., Zhu, J.Y., Zhou, T., Efros, A.A.: Image-to-image translation with
  conditional adversarial networks. In: Proceedings of the IEEE conference on
  computer vision and pattern recognition. pp. 1125--1134 (2017)

\bibitem{johnson2017clevr}
Johnson, J., Hariharan, B., Van Der~Maaten, L., Fei-Fei, L., Lawrence~Zitnick,
  C., Girshick, R.: Clevr: A diagnostic dataset for compositional language and
  elementary visual reasoning. In: Proceedings of the IEEE conference on
  computer vision and pattern recognition. pp. 2901--2910 (2017)

\bibitem{kang2020contragan}
Kang, M., Park, J.: Contragan: Contrastive learning for conditional image
  generation. In: NeurIPS (2020)

\bibitem{karras2019style}
Karras, T., Laine, S., Aila, T.: A style-based generator architecture for
  generative adversarial networks. In: Proceedings of the IEEE/CVF conference
  on computer vision and pattern recognition. pp. 4401--4410 (2019)

\bibitem{kim2017deepstory}
Kim, K.M., Heo, M.O., Choi, S.H., Zhang, B.T.: Deepstory: video story qa by
  deep embedded memory networks. In: Proceedings of the 26th International
  Joint Conference on Artificial Intelligence. pp. 2016--2022 (2017)

\bibitem{lei2020mart}
Lei, J., Wang, L., Shen, Y., Yu, D., Berg, T., Bansal, M.: Mart:
  Memory-augmented recurrent transformer for coherent video paragraph
  captioning. In: Proceedings of the 58th Annual Meeting of the Association for
  Computational Linguistics. pp. 2603--2614 (2020)

\bibitem{lester2021power}
Lester, B., Al-Rfou, R., Constant, N.: The power of scale for
  parameter-efficient prompt tuning. In: Proceedings of the 2021 Conference on
  Empirical Methods in Natural Language Processing. pp. 3045--3059 (2021)

\bibitem{lewis2020bart}
Lewis, M., Liu, Y., Goyal, N., Ghazvininejad, M., Mohamed, A., Levy, O.,
  Stoyanov, V., Zettlemoyer, L.: Bart: Denoising sequence-to-sequence
  pre-training for natural language generation, translation, and comprehension.
  In: Proceedings of the 58th Annual Meeting of the Association for
  Computational Linguistics. pp. 7871--7880 (2020)

\bibitem{LI2020102956}
Li, C., Kong, L., Zhou, Z.: Improved-storygan for sequential images
  visualization. Journal of Visual Communication and Image Representation
  \textbf{73},  102956 (2020).
  \doi{https://doi.org/10.1016/j.jvcir.2020.102956},
  \url{http://www.sciencedirect.com/science/article/pii/S1047320320301826}

\bibitem{li2021prefix}
Li, X.L., Liang, P.: Prefix-tuning: Optimizing continuous prompts for
  generation. In: Proceedings of the 59th Annual Meeting of the Association for
  Computational Linguistics and the 11th International Joint Conference on
  Natural Language Processing (Volume 1: Long Papers). pp. 4582--4597 (2021)

\bibitem{li2019storygan}
Li, Y., Gan, Z., Shen, Y., Liu, J., Cheng, Y., Wu, Y., Carin, L., Carlson, D.,
  Gao, J.: Storygan: A sequential conditional gan for story visualization. In:
  Proceedings of the IEEE Conference on CVPR. pp. 6329--6338 (2019)

\bibitem{liang2019cpgan}
Liang, J., Pei, W., Lu, F.: Cpgan: full-spectrum content-parsing generative
  adversarial networks for text-to-image synthesis. arXiv preprint
  arXiv:1912.08562  (2019)

\bibitem{liu2019roberta}
Liu, Y., Ott, M., Goyal, N., Du, J., Joshi, M., Chen, D., Levy, O., Lewis, M.,
  Zettlemoyer, L., Stoyanov, V.: Roberta: A robustly optimized bert pretraining
  approach. arXiv preprint arXiv:1907.11692  (2019)

\bibitem{mahabadi2021parameter}
Mahabadi, R.K., Ruder, S., Dehghani, M., Henderson, J.: Parameter-efficient
  multi-task fine-tuning for transformers via shared hypernetworks. In:
  Proceedings of the 59th Annual Meeting of the Association for Computational
  Linguistics and the 11th International Joint Conference on Natural Language
  Processing (Volume 1: Long Papers). pp. 565--576 (2021)

\bibitem{maharana2021integrating}
Maharana, A., Bansal, M.: Integrating visuospatial, linguistic, and commonsense
  structure into story visualization. In: Proceedings of the 2021 Conference on
  Empirical Methods in Natural Language Processing. pp. 6772--6786 (2021)

\bibitem{maharana2021improving}
Maharana, A., Hannan, D., Bansal, M.: Improving generation and evaluation of
  visual stories via semantic consistency. In: Proceedings of the 2021
  Conference of the North American Chapter of the Association for Computational
  Linguistics: Human Language Technologies. pp. 2427--2442 (2021)

\bibitem{mao2022unipelt}
Mao, Y., Mathias, L., Hou, R., Almahairi, A., Ma, H., Han, J., Yih, S., Khabsa,
  M.: Unipelt: A unified framework for parameter-efficient language model
  tuning. In: Proceedings of the 60th Annual Meeting of the Association for
  Computational Linguistics (Volume 1: Long Papers). pp. 6253--6264 (2022)

\bibitem{mitchell2019model}
Mitchell, M., Wu, S., Zaldivar, A., Barnes, P., Vasserman, L., Hutchinson, B.,
  Spitzer, E., Raji, I.D., Gebru, T.: Model cards for model reporting. In:
  Proceedings of the conference on fairness, accountability, and transparency.
  pp. 220--229 (2019)

\bibitem{van2016conditional}
Van~den Oord, A., Kalchbrenner, N., Espeholt, L., Vinyals, O., Graves, A.,
  et~al.: Conditional image generation with pixelcnn decoders. Advances in
  neural information processing systems  \textbf{29} (2016)

\bibitem{qiao2019mirrorgan}
Qiao, T., Zhang, J., Xu, D., Tao, D.: Mirrorgan: Learning text-to-image
  generation by redescription. In: Proceedings of the IEEE/CVF Conference on
  Computer Vision and Pattern Recognition. pp. 1505--1514 (2019)

\bibitem{radford2021learning}
Radford, A., Kim, J.W., Hallacy, C., Ramesh, A., Goh, G., Agarwal, S., Sastry,
  G., Askell, A., Mishkin, P., Clark, J., et~al.: Learning transferable visual
  models from natural language supervision. In: International Conference on
  Machine Learning. pp. 8748--8763. PMLR (2021)

\bibitem{ramesh2021zero}
Ramesh, A., Pavlov, M., Goh, G., Gray, S., Voss, C., Radford, A., Chen, M.,
  Sutskever, I.: Zero-shot text-to-image generation. In: International
  Conference on Machine Learning. pp. 8821--8831. PMLR (2021)

\bibitem{yolov3}
Redmon, J., Farhadi, A.: Yolov3: An incremental improvement. arXiv  (2018)

\bibitem{rennie2017self}
Rennie, S.J., Marcheret, E., Mroueh, Y., Ross, J., Goel, V.: Self-critical
  sequence training for image captioning. In: Proceedings of the IEEE
  conference on computer vision and pattern recognition. pp. 7008--7024 (2017)

\bibitem{sanh2019distil}
Sanh, V., Debut, L., Chaumond, J., Wolf, T.: Distilbert, a distilled version of
  bert: smaller, faster, cheaper and lighter. In: 5th Workshop on Energy
  Efficient Machine Learning and Cognitive Computing (NeurIPS) (2019)

\bibitem{sharma2018conceptual}
Sharma, P., Ding, N., Goodman, S., Soricut, R.: Conceptual captions: A cleaned,
  hypernymed, image alt-text dataset for automatic image captioning. In:
  Proceedings of the 56th Annual Meeting of the Association for Computational
  Linguistics (Volume 1: Long Papers). pp. 2556--2565 (2018)

\bibitem{song2020character}
Song, Y.Z., Rui~Tam, Z., Chen, H.J., Lu, H.H., Shuai, H.H.:
  Character-preserving coherent story visualization. In: European Conference on
  Computer Vision. pp. 18--33. Springer (2020)

\bibitem{song2020CPCSV}
Song, Y.Z., Tam, Z.R., Chen, H.J., Lu, H.H., Shuai, H.H.: Character-preserving
  coherent story visualization. In: Proceedings of the European Conference on
  Computer Vision (ECCV) (2020)

\bibitem{sung2022vl}
Sung, Y.L., Cho, J., Bansal, M.: Vl-adapter: Parameter-efficient transfer
  learning for vision-and-language tasks. In: Proceedings of the IEEE/CVF
  Conference on Computer Vision and Pattern Recognition. pp. 5227--5237 (2022)

\bibitem{szHucs2022modular}
Sz{\H{u}}cs, G., Al-Shouha, M.: Modular storygan with background and theme
  awareness for story visualization. In: International Conference on Pattern
  Recognition and Artificial Intelligence. pp. 275--286. Springer (2022)

\bibitem{thomee2016yfcc100m}
Thomee, B., Shamma, D.A., Friedland, G., Elizalde, B., Ni, K., Poland, D.,
  Borth, D., Li, L.J.: Yfcc100m: The new data in multimedia research.
  Communications of the ACM  \textbf{59}(2),  64--73 (2016)

\bibitem{van2017neural}
Van Den~Oord, A., Vinyals, O., et~al.: Neural discrete representation learning.
  Advances in neural information processing systems  \textbf{30} (2017)

\bibitem{xu2018attngan}
Xu, T., Zhang, P., Huang, Q., Zhang, H., Gan, Z., Huang, X., He, X.: Attngan:
  Fine-grained text to image generation with attentional generative adversarial
  networks. In: Proceedings of the IEEE conference on computer vision and
  pattern recognition. pp. 1316--1324 (2018)

\bibitem{yan2021videogpt}
Yan, W., Zhang, Y., Abbeel, P., Srinivas, A.: Videogpt: Video generation using
  vq-vae and transformers. arXiv preprint arXiv:2104.10157  (2021)

\bibitem{yin2019semantics}
Yin, G., Liu, B., Sheng, L., Yu, N., Wang, X., Shao, J.: Semantics
  disentangling for text-to-image generation. In: Proceedings of the IEEE/CVF
  conference on computer vision and pattern recognition. pp. 2327--2336 (2019)

\bibitem{yu2018generative}
Yu, J., Lin, Z., Yang, J., Shen, X., Lu, X., Huang, T.S.: Generative image
  inpainting with contextual attention. In: Proceedings of the IEEE conference
  on computer vision and pattern recognition. pp. 5505--5514 (2018)

\bibitem{zaken2022bitfit}
Zaken, E.B., Goldberg, Y., Ravfogel, S.: Bitfit: Simple parameter-efficient
  fine-tuning for transformer-based masked language-models. In: Proceedings of
  the 60th Annual Meeting of the Association for Computational Linguistics
  (Volume 2: Short Papers). pp.~1--9 (2022)

\bibitem{zeng2019pororogan}
Zeng, G., Li, Z., Zhang, Y.: Pororogan: An improved story visualization model
  on pororo-sv dataset. In: Proceedings of the 2019 3rd International
  Conference on Computer Science and Artificial Intelligence. pp. 155--159
  (2019)

\bibitem{zhang2021cross}
Zhang, H., Koh, J.Y., Baldridge, J., Lee, H., Yang, Y.: Cross-modal contrastive
  learning for text-to-image generation. In: Proceedings of the IEEE/CVF
  Conference on Computer Vision and Pattern Recognition. pp. 833--842 (2021)

\bibitem{han2017stackgan}
Zhang, H., Xu, T., Li, H., Zhang, S., Wang, X., Huang, X., Metaxas, D.:
  Stackgan: Text to photo-realistic image synthesis with stacked generative
  adversarial networks. In: {ICCV} (2017)

\bibitem{zhu2019dm}
Zhu, M., Pan, P., Chen, W., Yang, Y.: Dm-gan: Dynamic memory generative
  adversarial networks for text-to-image synthesis. In: Proceedings of the
  IEEE/CVF Conference on Computer Vision and Pattern Recognition. pp.
  5802--5810 (2019)

\end{thebibliography}

\appendix{}

\section{Background}
In this section, we give a brief introduction to the original story visualization task and auto-regressive transformers for text-to-image synthesis.

\subsection{Story Visualization}
\label{sec:sviz}
Given a sequence of sentences $S=[s_1, s_2, ..., s_T]$ forming a narrative, story visualization is the task of generating a corresponding sequence of images $\hat{X} = [\hat{x}_1, \hat{x}_2, ..., \hat{x}_T]$, following \cite{li2019storygan}. The sentences form a coherent story with recurring plot and characters. The generative model for this task has two main modules: story encoder and image generator. The sentence encoder $E_{caption}(.)$ takes word embeddings $\{w_{ik}\}$ for sentence $s_k$ at each timestep $k$ and generates contextualized embeddings $\{c_{ik}\}$. These embeddings are then used to generate the corresponding images. The terms \textit{caption} and \textit{sentence} are used interchangeably throughout the paper.

\subsection{Pretrained Text-to-Image Synthesis Models (DALL-E)}
The DALL-E model introduced in \cite{ramesh2021zero} is a text-to-image synthesis pipeline which comprises of a discrete variational autoencoder (dVAE) in the first stage and an autoregressive transformer in the second stage:

\paragraph{Stage 1.} The Vector Quantized Variational Autoencoder (VQVAE) \cite{van2016conditional} consists of an encoder that learns to map high dimensional input data ($x$) to a discretized latent space, and a decoder that reconstructs $x$ from the quantized encodings $x^{q}$. The model is trained using the reconstruction loss and commitment loss \cite{van2017neural}. In DALL-E, the VQVAE is trained to transform RGB image into a small 2D grid of image tokens, where each token can assume a discrete value from a codebook of predefined length.

\paragraph{Stage 2.} The VQVAE encoder from Stage 1 is used to infer the grid of discretized image tokens which is flattened and concatenated with the input text tokens, and an autoregressive transformer is used to model the joint distribution over the text and image tokens. For a given text input $s$ and target image $x$, these models learn the distribution of image tokens $p(x)$ as,
\begin{equation}
    p(x) = \prod_{i=1}^{d} p(x_{i}|x_{i<i};s)
|x<i)
\end{equation}
The models are composed of stacked multi-head self-attention layers with causal masking and are optimized via maximum likelihood. Each self-attention block is followed by a MLP feedforward layer, as per the standard design of transformers. The prediction of image tokens at each time step is influenced by the text tokens and previously predicted image tokens via the self-attention layer.

Using this framework, DALL-E obtains impressive, state-of-the-art results on a variety of text-to-image tasks by leveraging large-scale pre-training on multimodal datasets.

\section{Additional Method Details} \label{sec:modeling_appendix}
In this section, we provide additional details about the \sdalle{} and \sgan{} models.

\subsection{\sdalle{}}

\paragraph{Retro Cross-Attention Layer Density.} We experiment with different densities of cross-attention layers in our implementation of \sdalle{}. In the densest variation, we introduce the retro layer in every self-attention block of minDALL-E, effectively increasing the number of parameters in the model by nearly 60\%. We vary the density of the retro layer for one in every 1-5 self-attention block(s), and run experiments for each of these variations. Our best model has a density of one retro layer in every third self-attention block.

\paragraph{Objective.} Following the original DALL-E implementation \cite{ramesh2021zero}, the \sdalle{} model is trained on a combination of text loss and image loss. The losses are cross-entropy losses for the respective modalities, and the combined objective is,
\begin{equation*}
    \mathcal{L} = -\sum_{i=1}^{N_{text}}t_{i}log(p(t_{i}) -\sum_{i=1}^{N_{img}}m_{i}log(p(m_{i}) 
\end{equation*}
where $N_{text}$ and $N_{img}$ are the caption lengths and image sequence lengths, set to 64 and 256 in our model respectively.

\subsection{\sgan{}}
\sgan{} follows the general framework of the StoryGAN model \cite{li2019storygan} i.e., it is composed of a recurrent text encoder, an image generation module, and two discriminators - image and story discriminator. We modify this framework to accept the source frame as input for the story continuation task, and use it for improving the generation of target frames. Our \sgan{} model is implemented as follows:

\paragraph{Pre-trained Language Model Encoder.}
In the current state-of-the-art story visualization models \cite{maharana2021improving}, recurrent transformer-based text encoders like MART \cite{lei2020mart} and MARTT \cite{maharana2021integrating} are learnt from scratch for encoding the captions. However, while the memory module contains information about prior captions, there is no way for the current caption to directly attend to words in prior or subsequent captions. This is crucial in a story where causality plays such a large role, e.g., which characters need to appear in the scene, even if they don't appear in the current caption, has there been any modifications to the background that need to appear in the current scene, etc. Furthermore, general world knowledge is crucial for successfully generating unseen stories in our datasets, which is possible with pretrained knowledge. Therefore, we propose using a pretrained language model (such as RoBERTa \cite{liu2019roberta} or CLIP text encoder \cite{radford2021learning}) as the caption encoder. These models are pretrained on large unimodal or multimodal datasets of language; their latent knowledge of the world is of great utility for understanding the semantic concepts present in input captions. For the RoBERTa encoder \cite{liu2019roberta}, to ensure that the model has access to all captions, we append the captions together and feed all of them into each timestep. We use a special token to denote which caption is currently being generated. The representation from the first token $h_{0}$ is used as the caption representation. For the CLIP encoder \cite{radford2021learning}, we add an additional self-attention block that takes the caption representation for each timestep and produces the contextualized representations that have been computed by attending to all other timesteps.

\begin{figure*}[t]
    \centering
    \includegraphics[width=0.9\textwidth]{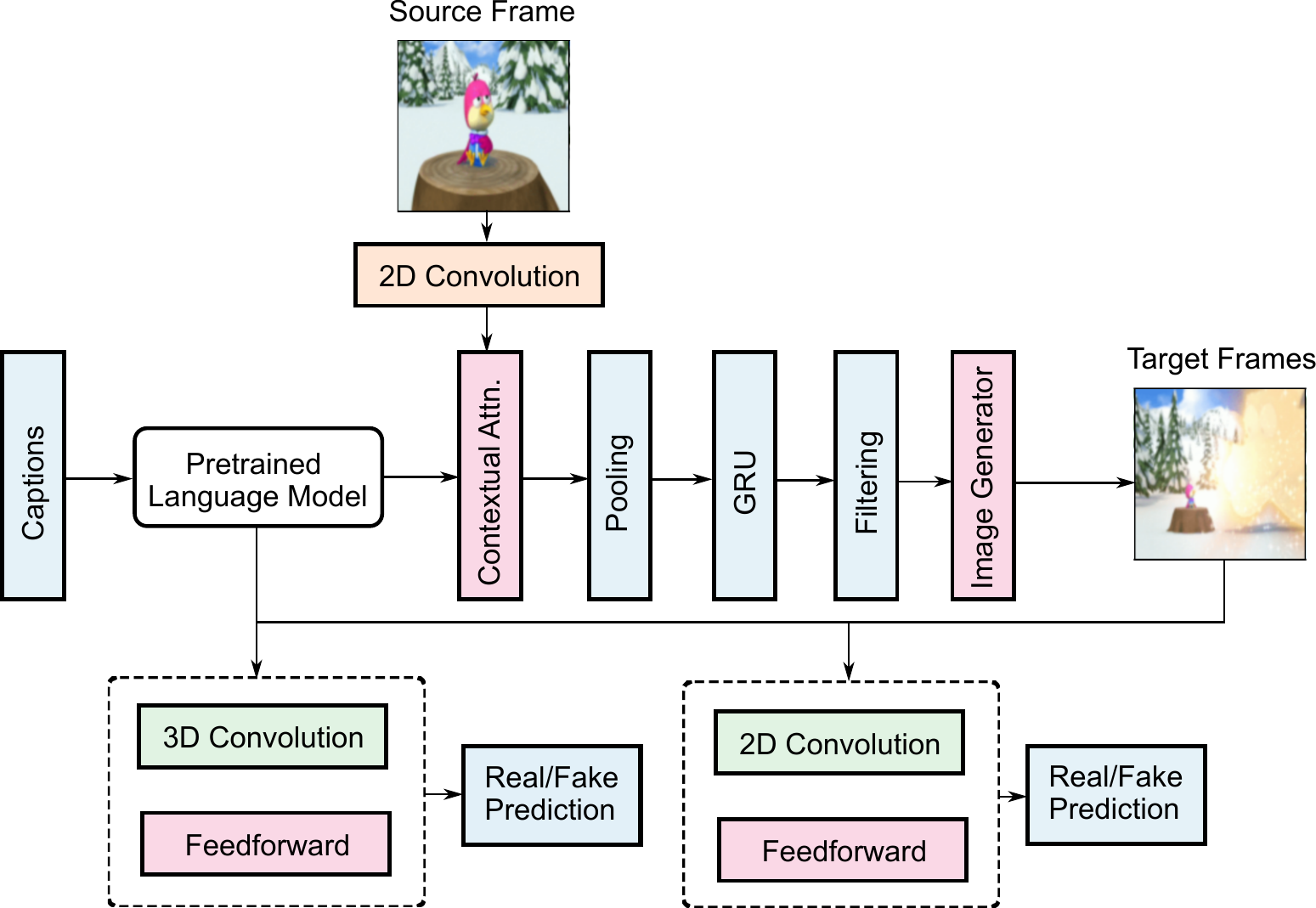}
    \caption{Illustration of our \sgan{} architecture. The captions are first encoded using a pretrained language model to produce contextualized representations. These representations are sent to a contextual attention module along with the source frame, and the resulting representation is sent to the image generator. The generated frames are sent to a story and image discriminator, and the corresponding cross-entropy losses for detection real/fake images are used to train the \sgan{} model.}
    \label{fig:model_storygan}
\end{figure*}

\paragraph{Contextual Attention.}
We then combine the story representation with the image embeddings of the first frame of the image sequence using contextual attention. First, we reshape the story representation as a 2D matrix and extract $3\times3$ patches $\{t_{x,y}\}$ as convolutional filters. Then, we match them against potential patches from the source frame $\{s_{x',y'}\}$ by measuring the normalized inner product as,
\begin{equation}
    p_{x,y,x',y'} = \langle \frac{s_{x,y}}{||s_{x,y}||},\frac{t_{x',y'}}{||t_{x',y'}||}\rangle
\end{equation}
where $p_{x,y,x,y'}$ represents the similarity between the patch centered in target frame ($x,y$) and source frame ($x',y'$). We compute the similarity score for all dimensions along ($x',y'$) for the patch in target frame ($x,y$) and find the best match from the softmax-scaled similarity scores. \cite{yu2018generative} implement this efficiently using convolution and channel-wise softmax; we use their implementation in our \sgan{} model. The extracted patches are used as deconvolutional filters and added to the target frame $s$. The resulting representation is fed through a generator module which processes each caption and produces an image. We use the generator module outlined in \cite{li2019storygan}.
\paragraph{Discriminators.}
Finally, the loss is computed for the generated image sequence. There are 3 different components that provide the loss for the model. The first is a story discriminator, which takes all of the generated images and uses 3D convolution to create a single representation and then makes a prediction as to whether the generated story is real or fake. Additionally, there is an image discriminator, which performs the same function but only focuses on individual images. Finally, the model is trained end-to-end using the objective function:
\begin{equation*}
    \min_{\theta_{G}} \max_{\theta_{I},\theta_{S}} \> \mathcal{L}_{KL} + \mathcal{L}_{img} + \mathcal{L}_{story}
\end{equation*}
where $\theta_{G}$, $\theta_{I}$ and $\theta_{S}$ denote the parameters of the text encoder + generator, and image and story discriminator respectively. $\mathcal{L}_{img}$ and $\mathcal{L}_{story}$ are cross-entropy losses for classifying ground truth and synthetic images into real and fake categories respectively. $\mathcal{L}_{KL}$ is the Kullback-Leibler (KL) divergence between the learned distribution $h_0$ and the standard Gaussian distribution, to enforce smoothness over the conditional manifold in latent semantic space \cite{li2019storygan}. During inference, the trained weights $\theta_{G}$ are used to generate a visual story for a given input of captions.

\section{Dataset Construction} \label{sec:datasets_arxiv}

We propose the new dataset DiDeMoSV, which is derived from the Didemo dataset \cite{hendricks17iccv}. Below, we present details about collection and cleaning of the dataset.

\subsection{Dataset Construction}
Prior work in story visualization has repurposed datasets from other tasks. We follow this trend and repurpose video captioning datasets in our work. Story visualization and video captioning share many components. In video captioning, an agent must produce a caption, or series of captions, that describe the content of a video. Story visualization can be thought of as video captioning in reverse, where frames are generated based on the captions. However, simply reversing the direction of the task is not sufficient in this case because the other difference between the two tasks is that story visualization has one frame per caption, whereas videos have many frames; a single caption is typically paired with a video time stamp, denoting which section of the video the caption aligns with. Therefore, to convert video captioning into story visualization, an appropriate method is needed to select which single frame should be used to represent the content of the caption.

We employ the self-critical image captioning model \cite{rennie2017self} for intelligently selecting the frame most aligned with the caption. Each of the clips that correspond to a caption is multiple seconds long. Not all of the frames will be equally aligned with the caption. Characters might be moving leaving blur effects, the scene might change a bit early or late in the clip, or there might be superfluous actions that occur. To initially shrink the number of frames that we must consider, we first sample frames at fixed intervals throughout the video. In the case of DiDeMoSV, we sample 10 frames. Each of the frames is then fed through the self critical image model and is ranked according to the sum of the log likelihood for each word in the caption being generated. We then use the top-ranked frame as the image for the given caption. The resulting image-caption sequence after this step is on average 4 frames long for DiDeMoSV. To maximize the amount of data that we have and make the task feasible, we split these image-caption sequences into a sequence of 3 frames. We use a sliding window approach to create these sequences, allowing for overlap between sequences. However, we also ensure that the train, val, and test splits contain separate videos. We then proceed with our image pre-processing steps.

The main pre-processing step that we explore is to convert the real-world images into cartoon images, to emphasize focus on the main characters of the image rather than the trivial details of the background. Rather than models focusing on making images realistic, we want them to focus on accurately representing the stories themselves in visual form. To cartoonize the images we use CartoonGAN \cite{chen2018cartoongan}. Each of the extracted frames is fed through this network and the resulting output is used in the final dataset.

\section{Experimental Details}
\paragraph{Pretrained Weights.}
While the VAE checkpoints for the original DALL-E model have been released, the transformer weights have not. We explored training the transformer component from scratch on our data, but found that it did not perform well. Therefore, we explored other publicly available efforts to reproduce DALL-E and settled on a popular open-source version minDALL-E which is composed of 1.3 billion parameters and trained on 14 million text-image pairs from the CC3M \cite{sharma2018conceptual} and CC12M \cite{changpinyo2021conceptual} datasets.\footnote{\url{https://github.com/kakaobrain/minDALL-E}}
minDALL-E uses the pretrained VQGAN-VAE \cite{esser2021taming} for discretizing image inputs. We adapt the pretrained model minDALL-E to StoryDALL-E and then prompt-tune/fine-tune the retro-fitted model on our target datasets.

We experiment with pretrained CLIP \cite{radford2021learning} (38M parameters) and distilBERT \cite{sanh2019distil} (110M parameters) text encoders for the LM-StoryGAN models. The CLIP image encoder is used to extract image embeddings for the source frame in the story continuation task. The universal sentence transformer \cite{cer2018universal} is used to extract sentence embeddings for captions, that are sent as input to the global story encoder in \sdalle{}.

\paragraph{Training Details.}
We conduct experiments in the story continuation setting, i.e., the models receive the first frame as input condition. The \sdalle{} and \sdallem{} models are trained for 5 epochs with learning rates of 1e-04 (AdamW, Cosine Scheduler) and 5e-04 (AdamW, Linear Decay Scheduler) for fine-tuning and prompt-tuning setups respectively. We use a cosine schedule with warmup from 0 in the first 750 training steps. The minimum learning rate is 0.1 times the maximum learning rate. Checkpoints are saved at the end of every epoch. In full-model finetuning settings, the pretrained weights are finetuned with a smaller learning rate of 1e-05. The LMStoryGAN models are trained for 120 epochs with learning rates 1e-04 and 1e-05 for the generator and discriminators respectively. Checkpoints are saved every 10 epochs. These models are trained on 1-2 A6000 GPUs.

For the publicly available demo, we have continued training the \sdalle{} and \sdallem{} models for up to 50 epochs, which takes up to 10 days on 2 A6000 GPUs and exhibits improved performance over the checkpoints reported in the paper. See the codebase for links to the demo and the checkpoints used therein.\footnote{\url{https://github.com/adymaharana/storydalle}}

\paragraph{Evaluation Metrics.}
We consider 3 automatic evaluation techniques. The first is FID score, which calculates the difference between the ground truth and generated images by computing the distance between two feature vectors. We follow prior work and use Inception-v3 as our image encoding model. 

Following \cite{li2019storygan} and \cite{maharana2021improving}, we also compute the character classification scores for the Pororo and Flintstones datasets, which are adapted from video QA datasets with recurring characters. We use the Inception-v3 models trained for character classification on these respective datasets for computing the F1 Score and frame accuracy (exact match). Since the DiDeMoSV dataset does not have recurring characters, we do not evaluate performance of our models on these datasets using character classification.

\section{Additional Results}
In this section, we present the results on validation sets of the three story continuation datasets discussed in Table~\ref{tab:test_fid} in main text.

\begin{table*}[t]
\small
\centering
\caption{\label{tab:val_fid}Results on the validation sets of PororoSV, FlintstonesSV and DiDeMoSV (DSV) datasets from various models. Scores are based on FID (lower is better), character classification F1 and frame accuracy (F-Acc.; higher is better) evaluations.}
\scalebox{0.9}{
\begin{tabular}{|c|c|c|c|c|c|c|c|}
\hline
\multirow{2}{*}{\textbf{Model}} & \multicolumn{3}{c|}{\textbf{ PororoSV }} & \multicolumn{3}{c|}{\textbf{ FlintstonesSV }} & \textbf{DSV}\\
\cline{2-8}
 & FID $\downarrow$&{Char-F1$\uparrow$}& F-Acc$\uparrow$& FID $\downarrow$&{Char-F1$\uparrow$}& F-Acc$\uparrow$ & FID$\downarrow$\\
\hline
\sgan{} (BERT) & 63.94 & 54.02 & 24.53 & 87.65 & 71.98 & 55.68 & 93.21 \\
\sgan{} (CLIP) & 65.13 & \textbf{54.83} & \textbf{25.29} & 87.02 & 72.30 & \textbf{59.35} & 93.26 \\
\sdalle{} (prompt) & 45.68 & 31.91 & 22.14 & 67.05 & 54.17 & 26.23 & 72.61 \\
\sdalle{} (finetuning) & \textbf{21.64} & \textbf40.28 & 20.94 & \textbf{28.37} & \textbf{74.28} & \textbf52.35 & \textbf{41.58} \\
\hline
\end{tabular}
}
\end{table*}

\paragraph{Validation Set Results.} We present results on the validation set of the three story continuation datasets discussed in main text i.e. PororoSV, FlintstonesSV and DiDeMoSV, in Table~\ref{tab:val_fid}. The fully-finetuned \sdalle{} model performs the best across all datasets in terms of FID score. The gains are seen in FID, due the high visual quality of the images generated by \sdalle{}. However, the character classification and frame accuracy scores for the \sdalle{} are close to those of \sgan{} for the FlintstonesSV dataset and relatively lower for the PororoSV dataset, in spite of being of better visual quality (as per manual analysis). This might be attributed to the fact that GAN-based models tend to generate some finer details of a character while sacrificing shape and form, which is recognized by character classification models as a faithful reconstruction. On the other hand, \sdalle{} models focus on shape and form and tend to blur other defining characteristics, which are appealing to human eyes but fail to be recognized by the classification model.

\begin{figure*}[t]
    \centering
    \includegraphics[width=0.9\textwidth]{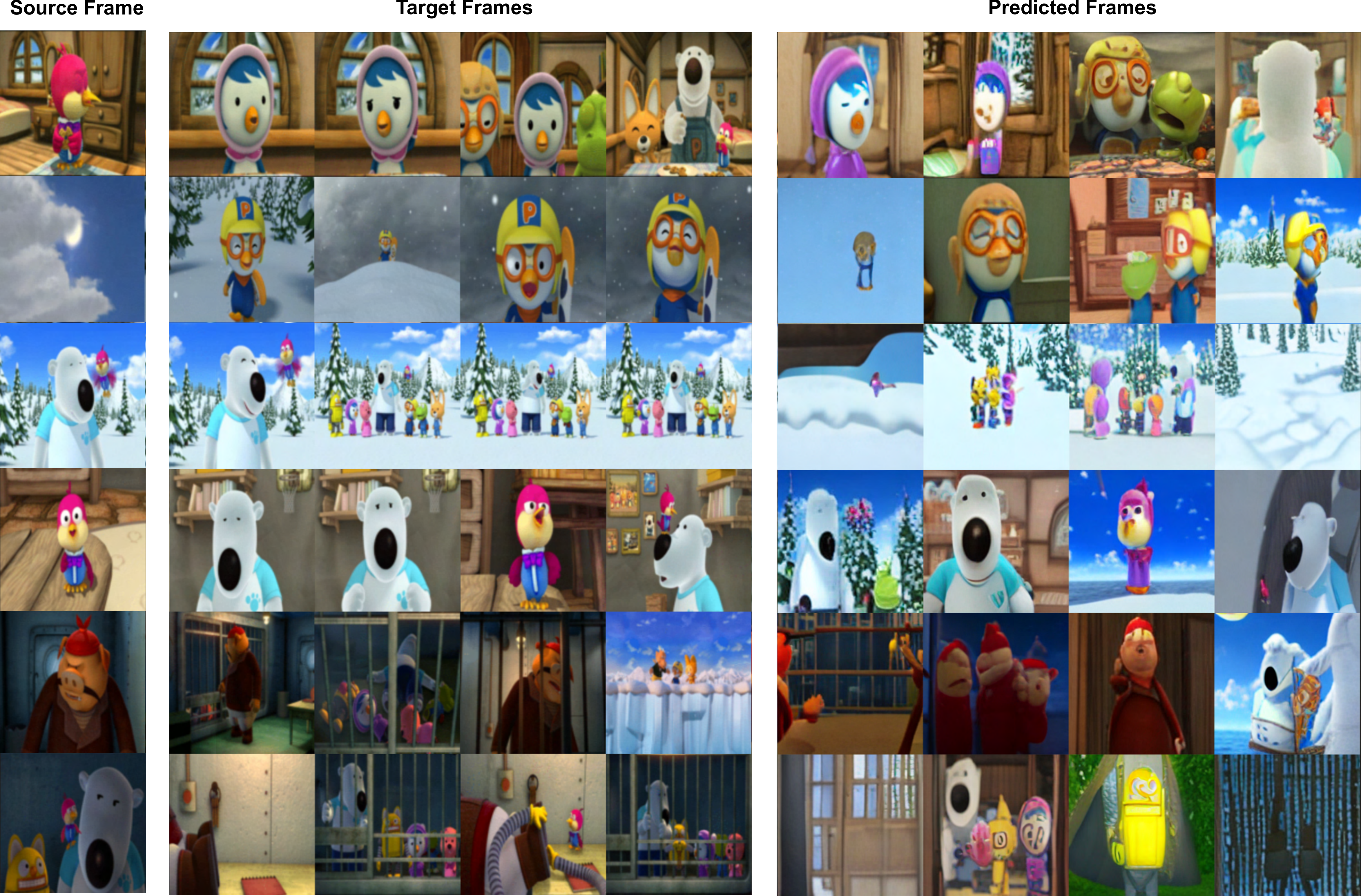}
    \caption{Generated samples from \sdalle{} for the PororoSV dataset.
    \label{fig:example_pororo}}
\end{figure*}

\begin{figure*}[t]
    \centering
    \includegraphics[width=0.9\textwidth]{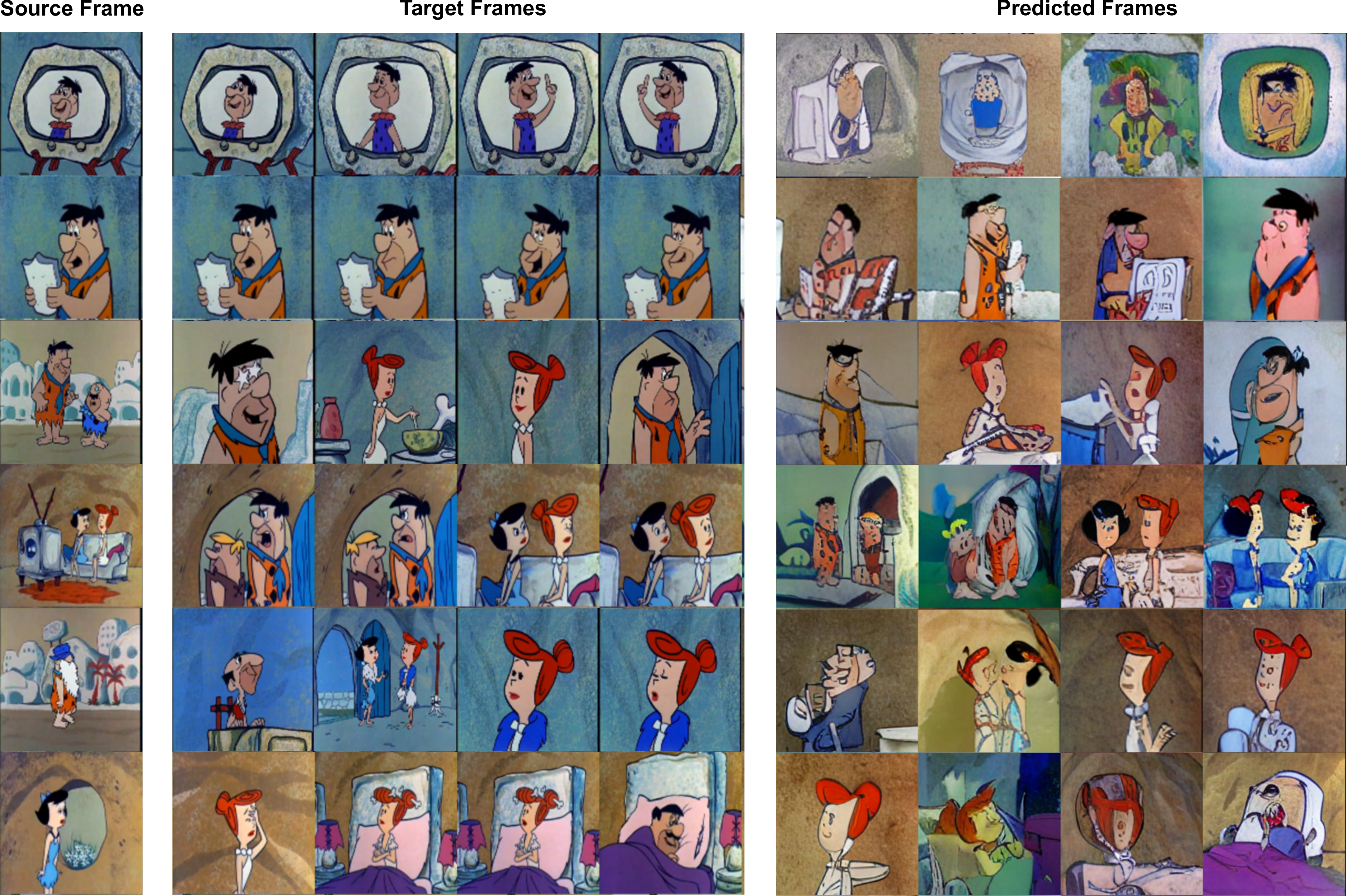}
    \caption{Generated samples from \sdalle{} for the FlintstonesSV dataset.
    \label{fig:example_flintstones}}
\end{figure*}

\begin{figure*}[t]
    \centering
    \includegraphics[width=0.9\textwidth]{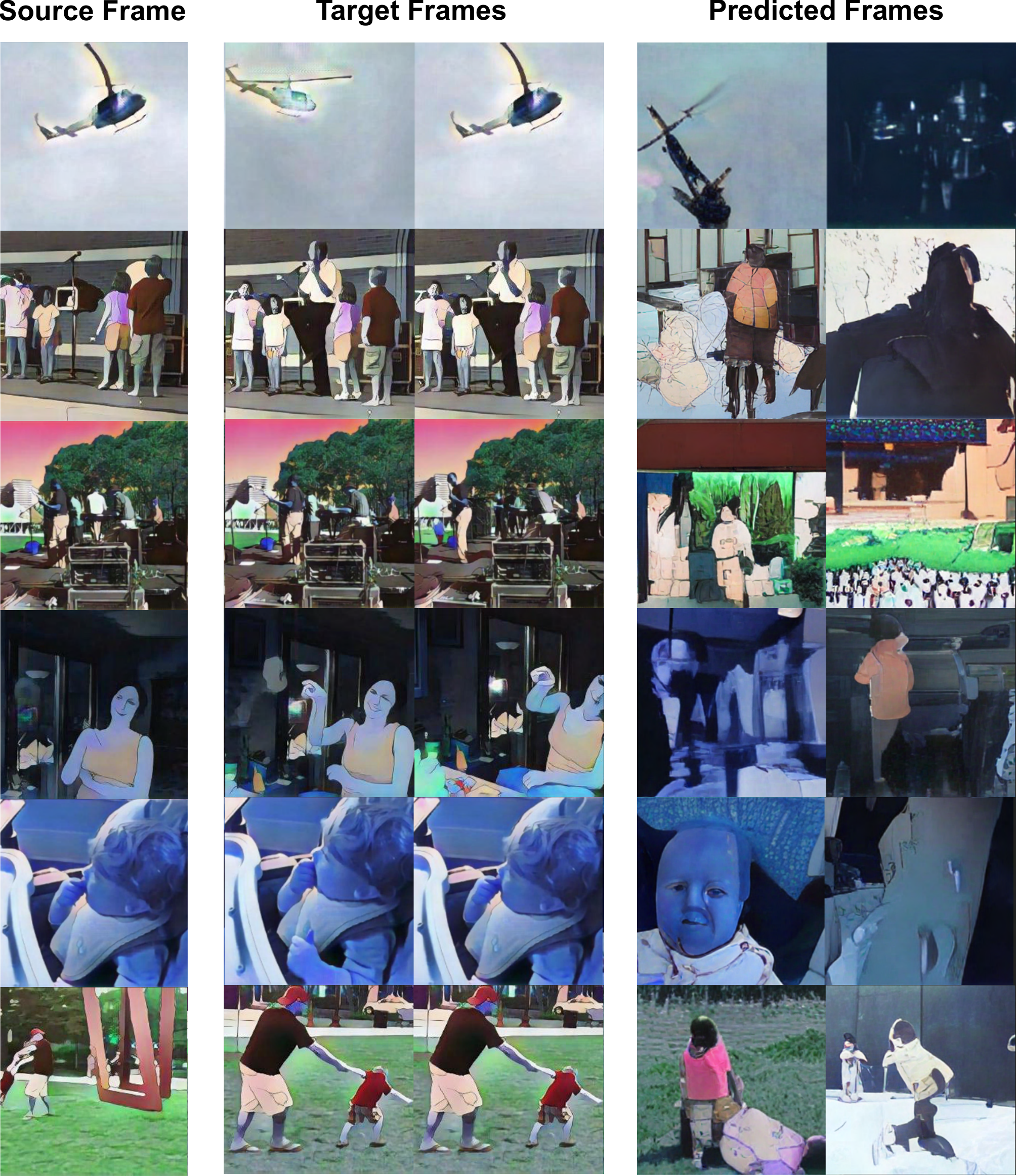}
    \caption{Generated samples from \sdalle{} for the DiDeMoSV dataset.
    \label{fig:example_didemo}}
\end{figure*}

Due to the higher resolution images generated by VQGAN-VAE \cite{esser2021taming}, the visual quality of images produced by \sdalle{} is highly preferred over predictions from the \sgan{} models. Similarly, the latent pretrained knowledge of DALL-E promotes generation of images that align well with the input captions, and results in higher wins for the \sdalle{} model. The \%wins and \%loss are nearly uniform for the attribute \textit{consistency} in this larger experiment, for the PororoSV and DiDeMoSV datasets. Predictions from the \sdalle{} model are found to be more consistent than those of \sgan{} for the FlintstonesSV dataset. See predictions from \sdalle{} for the PororoSV, FlintstonesSV and DiDeMoSV datasets in Figures~\ref{fig:example_pororo}, \ref{fig:example_flintstones} and \ref{fig:example_didemo} respectively.

\section{Additional Analysis}
In this section, we examine various aspects of the story continuation task, models and datasets. First, we demonstrate the advantages of the story continuation task over the story visualization task. Next, we calculate correlations between the source images and generated images from \sdalle{}, with and without condition, to demonstrate the utility of cross-attention layers. Finally, we discuss the semantic content of our proposed DiDeMoSV dataset.

\begin{figure*}
    \centering
    \includegraphics[width=0.8\textwidth]{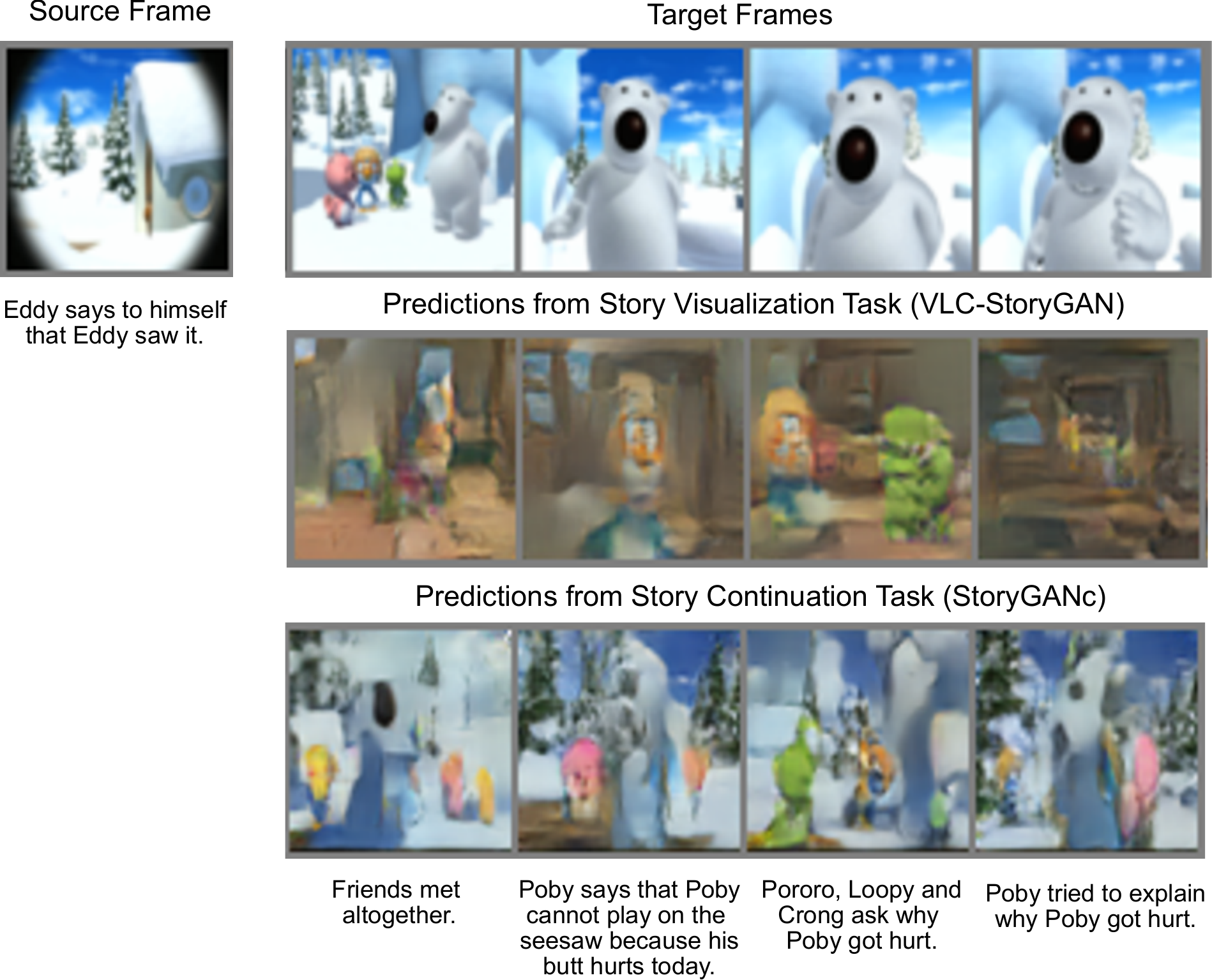}
    \caption{Comparison of predictions from state-of-the-art story visualization model VLCStoryGAN (middle) and our story continuation model \sgan{} (bottom) for a sample from the PororoSV dataset (top).}
    \label{fig:vizvcont}
\end{figure*}

\subsection{Story Visualization vs. Story Continuation}
In Figure~\ref{fig:vizvcont}, we present a comparison of predictions from the state-of-the-art story visualization model VLCStoryGAN \cite{maharana2021integrating} and our story continuation model \sgan{} for a sample from the test set of the PororoSV dataset. Story Visualization relies only on the input captions to generate the images from scratch. However, as discussed in Section 3.1 in the main text, the captions in story visualization datasets are short and do not contain information about the setting and background elements. As a result, the predictions from story visualization models rely on data seen in the training set to infer arbitrary visual elements. In Figure~\ref{fig:vizvcont}, the story takes place in a snowy field with trees (top), but the prediction from VLCStoryGAN (middle) depicts the story as taking place indoors. When the first frame is given as additional input to our model \sgan{} in the story continuation task, the models borrows the snowy fields from the source frame and creates the story within that setting (bottom). Hence, story continuation is a more realistic and practical version of story visualization that can enable significant progress in research and faster transfer of technology from research to real-world use cases. Our experiments and datasets demonstrate the utility of this task.

\subsection{Correlation between Source and Generated Images}

We also measure the cosine similarity between the source frames and the generated frames from \sdalle{}, with and without the retro-fitted cross-attention layer for conditioning on a source image, as a representation of the correlation between the two sets of images. We encode the images using the CLIP image encoder ViT-B/16 and report the mean and standard deviation of cosine similarity values for each dataset (see Table~\ref{tab:correlation}). We see up to 0.3 points increase in correlation between the source image and generated image for all three datasets with the use of the conditioning mechanism.

\begin{table}[h]
\centering
\caption{\label{tab:correlation} Mean and standard deviation of correlation between source image and generated images from \sdalle{} without and with conditioning on the source image.}
\begin{tabular}{ |c|c|c| } 
\hline
\textbf{Dataset} & \textbf{without condition} & \textbf{with condition}\\ 
\hline
PororoSV & 0.23 $\pm$ 0.04 & 0.26 $\pm$ 0.04 \\
FlintstonesSV  & 0.38 $\pm$ 0.05 & 0.41 $\pm$ 0.03\\
DiDeMoSV & 0.16 $\pm$ 0.04 & 0.19 $\pm$ 0.01\\
\hline
\end{tabular}
\end{table}

\subsection{Semantic Analysis of the DiDeMoSV dataset.}
\begin{figure*}
    \centering
    \includegraphics[width=0.9\textwidth]{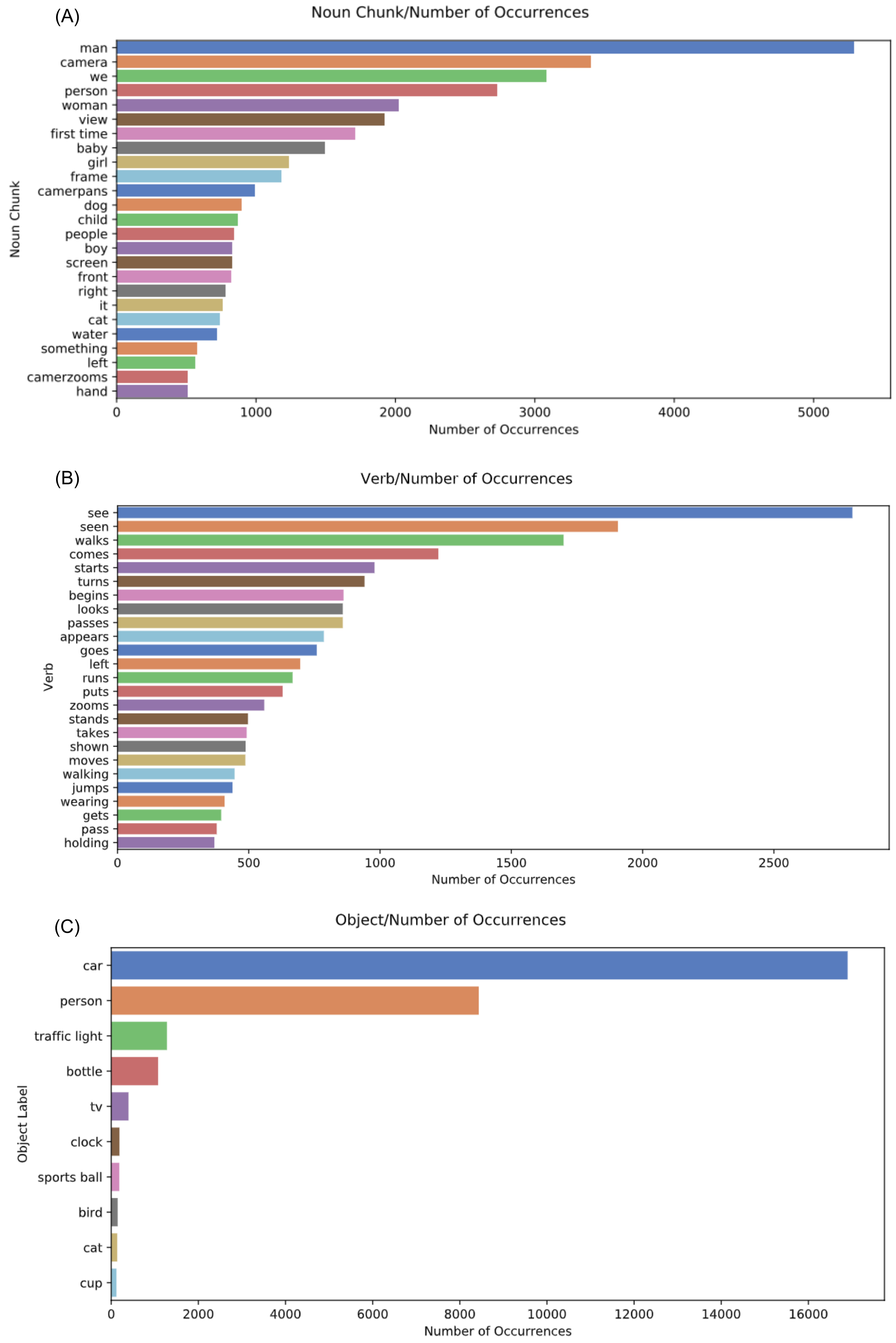}
    \caption{Plots for frequency of (A) noun chunks and (B) verbs in the captions and (C) objects in the frames of the DiDeMoSV dataset.}
    \label{fig:didemo_analysis}
\end{figure*}

Figure \ref{fig:didemo_analysis} contains counts for (A) noun chunks, (B) verbs and (C) object classes in DiDeMoSV. As discussed in Section \ref{sec:datasets_arxiv}, DiDeMoSV is collected from Flickr and the most common nouns indeed reflect this. Most of the captions are descriptive in that they describe the contents of the scene, the location of the objects/people in the scene, and the actions that are taking place in the scene. In DiDeMoSV, the focus is on the breadth of information that must be considered in the form of actions, objects, and settings.

The graph for the frequency of verbs across the captions in the DiDeMoSV dataset (see (B) in Figure~\ref{fig:didemo_analysis}) illustrates the complexity of the actions that are being undertaken by agents in the story. It can be seen that most of the actions are simplistic and related to movement, such as ``walks", ``comes", ``starts", ``turns", ``goes", etc. A lot of the verbs are also centered around vision, such as ``see", ``seen", and ``looks". While these words corroborate our prior insights reflecting the relative simplicity of the stories in DiDeMoSV, they also are crucial for understanding simplistic event chains. An understanding of these simple verbs and the way that they affect the story goes a long way towards facilitating story continuation, especially in the many settings of DiDeMoSV.

Part (C) in Figure ~\ref{fig:didemo_analysis} contains a breakdown of the objects that appear in the DiDeMoSV images. To generate these graphs, we use Yolov3 \cite{yolov3} to process each of the images in the respective datasets. The 'person' class is the dominant class in both datasets. This intuitively makes sense due to the initial data sources from which the respective video captioning datasets were constructed. Additionally, it matches the pattern that is observed in the caption noun analysis, where the nouns in both datasets are most frequently referring to people. However, we can also see that there are limitations of the Yolov3 model. There are frequently occurring nouns, such as `camera' in DiDeMoSV that are not able to appear in our image analysis because these do not have corresponding classes in the model. We use the default confidence threshold of 0.25 in the Yolo model, which generates predictions for only 76\% of DiDeMoSV images.

Our analysis demonstrates the diversity of the DiDeMoSV dataset, and showcases it as a challenging benchmark for the story continuation task, in addition to PororoSV and FlintstonesSV.

\end{document}